\documentclass[11pt]{article}

\PassOptionsToPackage{dvipsnames,table}{xcolor}
\PassOptionsToPackage{hyphens}{url}
\usepackage[preprint]{acl}
\usepackage{times}
\usepackage{latexsym}

\PassOptionsToPackage{dvipsnames,table}{xcolor}
\PassOptionsToPackage{hyphens}{url}

\usepackage[utf8]{inputenc}
\usepackage[T1]{fontenc}
\usepackage{microtype}
\usepackage{amsmath, amssymb, amsthm, amsfonts}
\usepackage{mathtools}
\usepackage{bm}

\makeatletter
\@ifpackageloaded{xcolor}{}{\usepackage[dvipsnames,table]{xcolor}}
\makeatother
\usepackage{graphicx}
\usepackage{placeins} % Keep the retention appendix figures with their section.
\usepackage{tikz}
\usetikzlibrary{patterns}
\usepackage{pgfplots}
\pgfplotsset{compat=1.18}
\usepackage{booktabs}
\usepackage{multirow}
\usepackage{tabularx}
\usepackage{array}
\usepackage{makecell}
\usepackage{siunitx}
\usepackage{arydshln}

\usepackage{algorithm}
\usepackage{algpseudocode}
\usepackage{listings}

\usepackage{url}
\makeatletter
\@ifpackageloaded{hyperref}{}{\usepackage{hyperref}}
\@ifpackageloaded{cleveref}{}{\usepackage[capitalize,noabbrev]{cleveref}}
\makeatother
\definecolor{aclLinkBlue}{HTML}{000099}
\hypersetup{
  colorlinks=true,
  linkcolor=aclLinkBlue,
  citecolor=aclLinkBlue,
  urlcolor=aclLinkBlue,
  pdftitle={WeAct: Dynamic Weight Switching as Agent Actions},
  pdfauthor={Rui Shi, Xunjian Yin, Jiacheng Sang, Zichuan Liu, Shuyan Zhou},
  pdfsubject={Research preprint},
}

\usepackage{ifthen}
\usepackage{xspace}

\newboolean{showcomments}
\setboolean{showcomments}{false}  % Hide author notes in the public preprint.

\newcommand{\makecomment}[3]{%
  \ifthenelse{\boolean{showcomments}}%
    {\textcolor{#1}{\textbf{[#2: #3]}}}{}%
}

\newcommand{\cut}[1]{\ifthenelse{\boolean{showcomments}}%
  {\textcolor{gray}{\sout{#1}}}{}}
\newcommand{\norm}[1]{\left\lVert#1\right\rVert}

\newcommand{\inner}[2]{\left\langle #1,\, #2 \right\rangle}

\newcommand{\method}{\texttt{WeAct}\xspace}  % rename in one place
\newcommand{\plan}{\texttt{LoRA-plan}\xspace}  % rename in one place
\newcommand{\search}{\texttt{LoRA-search}\xspace}  % rename in one place
\newcommand{\writing}{\texttt{LoRA-write}\xspace}  % rename in one place
\newcommand{\writingplus}{\texttt{LoRA-write++}\xspace}
\newcommand{\code}{\texttt{LoRA-code}\xspace}  % rename in one place
\newcommand{\dataset}[1]{\textsc{#1}\xspace}     % \dataset{ImageNet}
\newcommand{\model}[1]{\textsc{#1}\xspace}       % \model{GPT-4}

\newcommand{\best}[1]{\textbf{#1}}
\newcommand{\secondbest}[1]{\underline{#1}}
\theoremstyle{plain}
\newtheorem{theorem}{Theorem}

\theoremstyle{definition}

\theoremstyle{remark}
\newtheorem{remark}[theorem]{Remark}

\usepackage{array}
\usepackage{listings}
\usepackage{algorithm}
\usepackage{algpseudocode}
\usepackage{xspace}
\newcolumntype{C}[1]{>{\centering\arraybackslash}p{#1}}

\title{WeAct: Dynamic Weight Switching as Agent Actions}
\usepackage{latexml}

\iflatexml
\usepackage{authblk}
\makeatletter
\@ifundefined{ltx@authors@multiline}{}{\ltx@authors@multiline}
\makeatother
\author[1]{Rui Shi\textsuperscript{*\textdagger}}
\affil[1]{Carnegie Mellon University\\ City University of Hong Kong\\
  \href{mailto:ruishi@andrew.cmu.edu}{\texttt{ruishi@andrew.cmu.edu}}}
\author[2]{Xunjian Yin\textsuperscript{*}}
\affil[2]{Duke University\\
  \href{mailto:xunjian.yin@duke.edu}{\texttt{xunjian.yin@duke.edu}}}
\author[3]{Jiacheng Sang\textsuperscript{*}}
\affil[3]{Duke University\\
  \href{mailto:jiacheng.sang@duke.edu}{\texttt{jiacheng.sang@duke.edu}}}
\author[4]{Zichuan Liu}
\affil[4]{Carnegie Mellon University\\
  \href{mailto:zichuanl@andrew.cmu.edu}{\texttt{zichuanl@andrew.cmu.edu}}}
\author[5]{Shuyan Zhou\textsuperscript{\textdaggerdbl}}
\affil[5]{Duke University\\ Meta MSL\\
  \href{mailto:shuyan.zhou@duke.edu}{\texttt{shuyan.zhou@duke.edu}}}
\else
\author{
  Rui Shi\textsuperscript{1,2}\textsuperscript{*\textdagger}\quad
  Xunjian Yin\textsuperscript{3}\textsuperscript{*}\quad
  Jiacheng Sang\textsuperscript{3}\textsuperscript{*}\quad
  Zichuan Liu\textsuperscript{1}\quad
  Shuyan Zhou\textsuperscript{3,4}\textsuperscript{\textdaggerdbl}\\
  \textsuperscript{1}Carnegie Mellon University \quad \textsuperscript{2}City University of Hong Kong \quad
  \textsuperscript{3}Duke University \quad 
  \textsuperscript{4}Meta MSL \quad 
  \\
  \texttt{\{ruishi,zichuanl\}@andrew.cmu.edu}\\
  \texttt{\{xunjian.yin,jiacheng.sang,shuyan.zhou\}@duke.edu}
}
\fi

\begin{document}

\maketitle

\begingroup
  \renewcommand{\thefootnote}{\fnsymbol{footnote}}
  \footnotetext[1]{Equal contribution.}\footnotetext[2]{Work partly done at City University of Hong Kong.}\footnotetext[3]{Work done while at Duke University.}
\endgroup

\begin{abstract}

Long-horizon agents must learn heterogeneous capabilities from interleaved trajectories. Updating shared parameters to learn new capabilities can degrade previously acquired ones, resulting in catastrophic forgetting. As a trajectory unfolds, different stages may require different capabilities, yet conventional joint training and static adapter merging maintain a fixed parameter configuration throughout execution. We introduce \textbf{\method} (\textbf{We}ight-level Internal \textbf{Act}ions), which trains independent capability-specific LoRA adapters over a frozen backbone and adds adapter selection to the agent's action space. These internal actions use the model's native tool-calling interface for on-demand capability composition in a shared context. Across eight deep-research benchmarks, \model{Qwen3-32B} with \method improves average accuracy by 6.2\% relative to full fine-tuning counterparts despite using less compute, and by 27.5\% relative to the strongest static LoRA-merging method. \method{} also supports modular continual adaptation, by adding a new code computation adapter to expand the agent's capabilities and updating the existing writing adapter to improve report quality. Both updates leave the backbone and other adapters unchanged, supporting modular continual adaptation in agents. 
\end{abstract}

% Queue the single-column teaser for the first page's right-column top.
\suppressfloats[t]
\begin{figure}[t]
    \centering
    \includegraphics[width=\columnwidth]{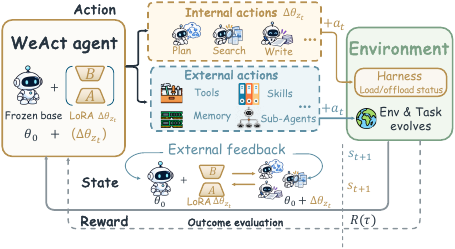}
    \caption{Internal actions extend the agent's action space. WeAct selects LoRA adapters internally and interacts with the environment through external actions.}
    \label{fig:weact-teaser}
\end{figure}

\section{Introduction}

Long-horizon agents plan and invoke tools, connect evidence across sources and reasoning steps, and retrieve and synthesize web information over many turns. The required capabilities vary across stages and must adapt as domains, tools, and output requirements evolve. Fine-tuning or reinforcement learning~\citep{websailorv2,stepdeepresearch}  typically optimize these capabilities in shared parameters: misaligned capability gradients can impair joint learning~\citep{dart}, and later updates can erode earlier gains~\citep{captrack2026}. Prompt-level skills~\citep{trace2skill} offer adaptation without changing weights, but their instructions consume context when loaded for execution.

Separate LoRA adapters isolate capability parameters, but their composition adds deployment requirements. MoLE~\citep{lora-moe} and LD-MoLE~\citep{ldmole} mix expert outputs through layerwise gates, requiring support for composition inside the forward pass. Static merging, from conflict reduction in TIES~\citep{tiesmerging} to alignment in KnOTS~\citep{knots}, combines updates before deployment. These methods improve composition, but use fixed merged weights across stages, which may weaken the specialization needed at each stage. Our key idea is to make independently maintained capabilities callable as agent actions. Extending ReAct's external actions~\citep{yao2023react}, we use \textbf{internal actions} to activate learned weights with on-demand flexibility.

We introduce \textbf{\method} (\textbf{We}ight-level Internal \textbf{Act}ions; Figure~\ref{fig:weact-teaser}), a general method for capability composition in tool-using agents, to learn heterogeneous capabilities for long-horizon tasks while preserving prior gains and invoking the capability needed at each stage. Each capability LoRA is trained over a frozen backbone on its own stages and exit actions, using preceding interactions as context. It learns to continue prior work, reason, call tools, and hand back control. At inference, our \texttt{load\_lora} and \texttt{offload\_lora} tools activate and release one adapter through the native tool-calling interface, with their descriptions and control exchanges recorded in context. The selected weights govern subsequent reasoning and tool use within the same history. \method-soft chooses capabilities from this history; \method-hard prescribes loading every adapter in the workflow for reliable invocation. Both use the same interface without a separately trained router. Adding or refreshing a capability updates only its adapter.

Experiments on \model{Qwen3-32B} test capability separation during learning and execution. Controlled diagnostics show that extra shared capacity does not prevent lost gains and that tested static merges reduce specialists' target-text fit (Section~\ref{sec:capability-separation}). Across eight deep-research benchmarks, \method-hard reaches 45.47\% average accuracy, a 6.2\% relative improvement over Full SFT, and both variants outperform all tested static merges on every benchmark (Section~\ref{sec:exp:main}). \method uses fewer trainable parameters and less nominal training compute than Full SFT while retaining the evaluated general capabilities. Code insertion expands the library with stable performance, and Writing refresh improves report quality on ResearchQA and DeepResearchBench (Section~\ref{sec:continual-adaptation}). These results suggest potential for modular continual adaptation through capability expansion and local refinement.

Our contributions are:
\begin{itemize}
  \item We present \method for independent capability training and coordinated invocation via native tools in a shared history.
   \item We distinguish learning interference from static fusion mismatch, motivating capability-aligned training and intact adapter selection.
  \item We demonstrate improved deep-research performance alongside capability insertion and local refinement, supporting continual learning through modular expansion and updates.
  \end{itemize}

\section{Method}
\label{sec:method}
\begin{figure*}[t]
    \centering
    \includegraphics[width=\linewidth]{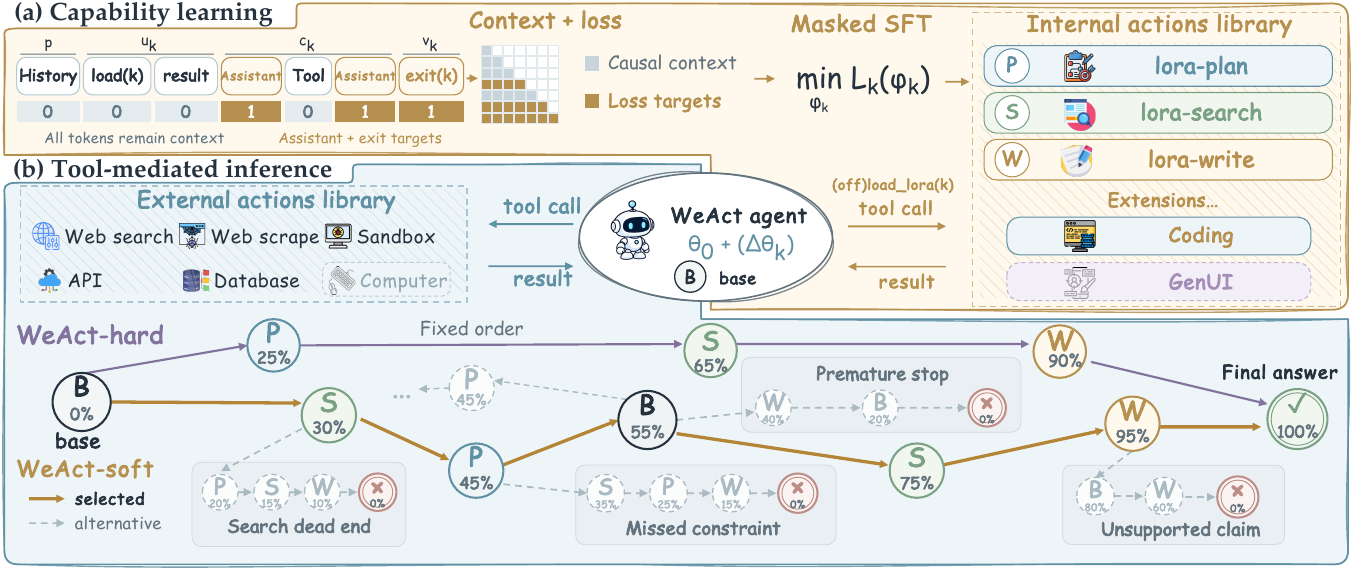}
    \caption{\method overview. (a) Capability learning builds the adapter library. (b) Tool-mediated inference switches adapters through internal tool calls and interacts with the environment through external tools. B/P/S/W nodes denote base/Plan/Search/Writing states; solid and dashed arrows indicate selected and alternative execution paths. }
    \label{fig:weact-overview}
\end{figure*}
\subsection{Preliminaries and Overview}
\label{sec:action-formalism}

Given a research query $q$, an agent gathers evidence and answers through successive reasoning and tool interactions. We train capability adapters independently and select them as the task unfolds. Let $\theta_0$ denote a frozen language model's parameters and $\{\Delta\theta_k\}_{k=1}^{K}$ a capability adapter library. Let $\mathcal M$ index the linear maps adapted by LoRA. For a map $\ell\in\mathcal M$ with input and output dimensions $d_{\mathrm{in},\ell}$ and $d_{\mathrm{out},\ell}$, LoRA~\citep{lora2022} gives
\begin{equation}
\label{eq:capability-lora}
\begin{gathered}
 W_{k,\ell}=W_{0,\ell}+\frac{\alpha_k}{r_k}B_{k,\ell}A_{k,\ell},\\
 A_{k,\ell}\in\mathbb R^{r_k\times d_{\mathrm{in},\ell}},\quad
 B_{k,\ell}\in\mathbb R^{d_{\mathrm{out},\ell}\times r_k}.
\end{gathered}
\end{equation}
Here $r_k$ and $\alpha_k$ are the rank and scaling coefficient. The trainable factors $\phi_k=\{A_{k,\ell},B_{k,\ell}\}_{\ell\in\mathcal M}$ determine $\Delta\theta_k=\Delta\theta_k(\phi_k)$; different capabilities may use different ranks. We write $\pi_k=\pi_{\theta_0+\Delta\theta_k}$ and set $\Delta\theta_\varnothing=0$ for base-only generation.

At step $t$, history $h_t$ retains instructions, model turns, and tool outputs. With at most one adapter active, $z_t\in\{\varnothing,1,\ldots,K\}$ selects the policy:
\begin{equation}
\label{eq:active-policy}
 a_t\sim\pi_{z_t}(\cdot\mid h_t)
       =\pi_{\theta_0+\Delta\theta_{z_t}}(\cdot\mid h_t).
\end{equation}

$a_t$ may contain reasoning or tool calls; internal tool calls change $z_t$ for subsequent generation.

Figure~\ref{fig:weact-overview} connects training and inference. During training, stage-specific assistant targets and exit calls supervise independent adapters over retained execution context. We group deep-research stages into planning, evidence search and verification, and writing, yielding \plan{}, \search{}, and \writing{}. These stages separate deciding what to do next, acquiring and checking external evidence, and synthesizing the answer. In exploratory annotation, finer splits were unstable: separating retrieval from verification or adding a state-recap stage let small rule changes move the same evidence operation between labels, and folding retrieval into planning made Plan absorb evidence-gathering steps (Appendix~\ref{app:segmentation-sensitivity}). We therefore keep evidence operations together in \search{}. During inference, external tools supply observations and internal actions select weights, with both exchanges extending the shared history. \textit{WeAct-hard} prescribes adapter invocation for reliable use; \textit{WeAct-soft} selects capabilities according to the evolving task state in that history.

\subsection{Capability Learning with Execution Context}
\label{sec:training}

Shared updates risk gradient interference and catastrophic forgetting. We isolate capability updates in separate adapters to prevent one capability from overwriting another's parameters. Consider shared trainable parameters $\omega$ and capability losses $\mathcal L_j^{\mathrm{shared}}$. With $g_i=\nabla_\omega\mathcal L_i^{\mathrm{shared}}(\omega)$ for $i\in\{j,k\}$, a gradient step for capability $k$ changes capability $j$'s loss to first order as
\begin{equation}
\label{eq:shared-update}
\begin{aligned}
 &\mathcal L_j^{\mathrm{shared}}(\omega-\eta g_k)
 -\mathcal L_j^{\mathrm{shared}}(\omega)\\
 &\qquad=-\eta\,g_j^\top g_k+O(\eta^2),
\end{aligned}
\end{equation}
for a locally smooth loss and step size $\eta$. The alignment and magnitude of these gradients determine the first-order interaction; Section~\ref{sec:capability-separation} measures the resulting retention effects.

We define capabilities by stage function, each potentially spanning several assistant/tool exchanges. \plan{} develops goals and action plans; \search{} covers evidence acquisition, filtering, and verification; \writing{} synthesizes evidence into an answer.

Each training sample retains preceding work so the module learns to continue an ongoing task. For capability $k$, let $p$ be the retained preceding history, $u_k$ the loading call and result, $c_k$ the stage's interleaved assistant/tool content, and $v_k$ its terminal assistant exit call. Concatenating these tokenized segments gives
\begin{equation}
\label{eq:training-sequence}
\begin{gathered}
 s=p\oplus u_k\oplus c_k\oplus v_k=(x_1,\ldots,x_N),\\
 m_{k,n}=\mathbf 1\{n\in\mathcal A_k(s)\}.
\end{gathered}
\end{equation}

The set $\mathcal A_k(s)$ contains the assistant-token positions used to supervise the current capability, including those corresponding to the exit call. The loss mask is zero for earlier assistant turns, the loading exchange, and tool observations. For the stage dataset $\mathcal D_k$, each module minimizes the objective:
\begin{equation}
\label{eq:masked-sft}
\begin{gathered}
 \mathcal L_k(\phi_k)
 =-\mathbb E_{s\sim\mathcal D_k}
       \sum_{n=1}^{N}m_{k,n}\log\pi_k(x_n\mid x_{<n}),\\
 \phi_k^\ast\in\arg\min_{\phi_k}\mathcal L_k(\phi_k).
\end{gathered}
\end{equation}
The mask selects prediction targets; all retained tokens remain causal context. In training and inference, \method{} retains the initial task and latest five user/tool messages, replacing older observation bodies with placeholders. We provide a complete training example in Appendix~\ref{app:training-example}.

For $j\ne k$, $\nabla_{\phi_j}\mathcal L_k=0$: optimizing capability $k$ leaves $\phi_j$ and $\theta_0$ unchanged, avoiding the shared-parameter interference in Eq.~\ref{eq:shared-update}. Refreshing a capability updates only its adapter factors, preserving the backbone and other modules' parameters without data replay, as in the Writing refresh in Section~\ref{sec:exp:writing-refresh}.

\subsection{Tool-Mediated Weight Actions}
\label{sec:architecture}

Separate adapters require a way to select capabilities as task needs change. We expose weight selection through native tool calls, reusing the model's learned tool selection and argument generation without a separately trained router. This interface fits existing tool harnesses and lets new capabilities enter through registry entries.

Our harness extends MiroFlow~\citep{miroflow} with two MCP tools, \texttt{load\_lora} and \texttt{offload\_lora}. Both take a registry alias, \texttt{lora\_name}, as their argument. Each call $a$ and result $r$ extend the history, $h\gets h\oplus a\oplus r$. After validation and successful execution, load/offload calls update the active index:
\begin{equation}
\label{eq:adapter-transition}
 z_t'=\begin{cases}
 k, & a_t=\operatorname{Load}(k),\ z_t=\varnothing,\\
 \varnothing, & a_t=\operatorname{Offload}(k),\ z_t=k,\\
 z_t, & \text{otherwise}.
 \end{cases}
\end{equation}
Thus switching from $i$ to $j$ releases $i$ before loading $j$. External, failed, rejected, and idempotent calls leave $z_t$ unchanged; reloading the active alias or offloading an empty state is a no-op. The harness checks aliases, lifecycle rules, and policy constraints, returning status, errors, or stage guidance. These changes to the weights governing subsequent generation are \emph{Weight-level Internal Actions}.

% A full-column algorithm respects the ACL two-column layout.
\begin{algorithm}[t]
\caption{\method inference.}
\label{alg:inference}
\fontsize{10}{12}\selectfont
\begin{algorithmic}[1]
\Require $q$, $\{\Delta\theta_k\}_{k=1}^{K}$, \method{}-hard/soft policy
\State $h\gets q$; $z\gets\varnothing$
\While{budget remains}
    \State $a\sim\pi_z(\cdot\mid h)$; $h\gets h\oplus a$
    \For{each tool call $c$ in $a$}
        \State $r\gets c()$; $h\gets h\oplus r$
        \If{$c=\texttt{load\_lora}(k)$}
            \State $z\gets k$
        \ElsIf{$c=\texttt{offload\_lora}(k)$}
            \State $z\gets\varnothing$
        \EndIf
    \EndFor
    \State Apply \method{}-hard/soft to $(h,z)$
    \State \textbf{if} answer ready and permitted \textbf{return} answer
\EndWhile
\State \Return budget-exhaustion result
\end{algorithmic}
\end{algorithm}

The policies differ in invocation timing. Under \method{}-soft, the base selects a module and the active module releases control, with revisits subject to coverage, repetition, and termination constraints. \textit{WeAct-hard} prescribes adapter use in the same agent, by default \plan{}$\to$\allowbreak\search{}$\to$\allowbreak\writing{}. The harness advances when the active stage emits an exit call, exhausts its budget, or reaches the no-tool threshold, recording release/load exchanges and handling termination or base-model fallback. Algorithm~\ref{alg:inference} outlines both.

Both control policies use the same execution mechanism: load and offload calls activate and release adapters that remain resident on the GPU, avoiding repeated host-to-device transfers. Weight changes require rebuilding the KV cache from the retained history under the new weights (Appendix~\ref{app:rq2-system}); answer-ready latency remains lower than Full SFT in the controlled evaluation (Section~\ref{sec:exp:training-efficiency}). The same tool interface also supports new capabilities: registering a new adapter with an alias and a capability description makes it available for selection, as in the Code insertion in Section~\ref{sec:exp:code-extension}.

\begingroup
\raggedbottom

\section{Understanding Capability Separation}
\label{sec:capability-separation}
\label{sec:exp:part1}

Learning heterogeneous capabilities through shared parameters, can erode earlier gains. Training separate LoRAs isolates their updates, but static fusion can alter each specialist's effects during execution. We use controlled retention and fusion diagnostics to examine these two sources of interference and motivate \method's capability-aligned parameter partitioning during learning and separate adapter selection during execution.

\subsection{Capability Interference and Retention}
\label{sec:exp:adapter-diagnostics}
\label{sec:exp:ablations}
\label{sec:exp:sequential-retention}

\begin{figure}[t]
\centering
\includegraphics[width=\linewidth]{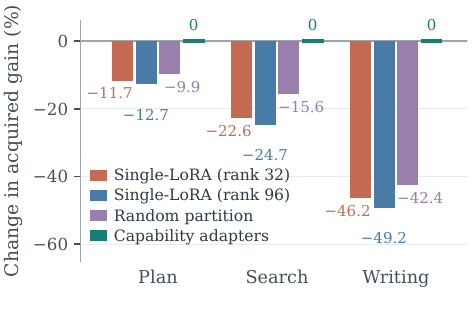}
\caption{Capability retention under sequential training. Detailed results are reported in Appendix Table~\ref{tab:capability-retention}.
}
\label{fig:capability-retention}
\end{figure}

We test whether shared updates erode earlier capabilities (Eq.~\ref{eq:shared-update}). Plan, Search, and Writing define successive trajectory stages; sequential training models capability-specific fine-tuning after initial SFT (Section~\ref{sec:continual-adaptation}), as in sequential continual LoRA learning~\citep{wang-etal-2025-see}.

On frozen \model{Qwen3-32B}, we train these capabilities in all six orders under identical data and schedules. Shared LoRAs of rank 32 or 96 match the active rank or total parameter count of three rank-32 adapters, testing whether extra capacity suffices. The partitioned adapters are assigned by capability, as in \method, or randomly per trajectory, testing whether separation suffices without capability alignment.

Capability-aligned adapters remain unchanged during later stages, providing a fixed reference for capability drift. We report teacher-forced top-1 content-token accuracy on 168 validation trajectories, with a separate 24-trajectory monitoring subset (one seed; Appendix~\ref{app:sequential-retention}); gradient diagnostics are in Appendix~\ref{app:rq1-metrics}.

% ======================================================================
% Main results table: benchmark-family columns and adaptation-family rows.
% ======================================================================
\begin{table*}[t]
\centering
% 9 pt text so every score fits with its standard error as a subscript.
\fontsize{9}{10.5}\selectfont
\providecommand{\tabse}[1]{\textsubscript{$\pm$#1}}
\setlength{\tabcolsep}{2pt}
\setlength{\aboverulesep}{1pt}
\setlength{\belowrulesep}{1pt}
\definecolor{oursband}{RGB}{228,243,255}
% Equal-width score columns; ordinary padding keeps the blue band continuous.
% The extra white gutter before Avg. is the only break in that band.
\newlength{\mainresultwidth}
\setlength{\mainresultwidth}{\dimexpr(\linewidth-82pt-36pt-3pt)/9\relax}
\begin{tabular}{@{}>{\raggedright\arraybackslash}p{82pt}*{8}{>{\centering\arraybackslash}p{\mainresultwidth}}@{\hspace{7pt}}>{\centering\arraybackslash}p{\mainresultwidth}@{}}
\toprule
\multicolumn{1}{c}{\textbf{Method}}
& \multicolumn{1}{c}{\textbf{General}}
& \multicolumn{3}{c}{\textbf{Multi-Hop QA}}
& \multicolumn{4}{c}{\textbf{Search}} & \\
\cmidrule(lr){2-2}\cmidrule(lr){3-5}\cmidrule(lr){6-9}
& \textbf{GAIA}
& \textbf{2Wiki}
& \makecell[tc]{\textbf{Bam-}\\\textbf{boogle}}
& \makebox[\linewidth][c]{\textbf{MuSiQue}}
& \makebox[\linewidth][c]{\makecell[tc]{\textbf{Web}\\\textbf{WalkerQA}}}
& \makecell[tc]{\textbf{xbench-}\\\textbf{DS}}
& \makecell[tc]{\textbf{Browse}\\\textbf{Comp}}
& \makebox[\linewidth][c]{\textbf{FRAMES}}
& \textbf{Avg.}
\\
\midrule
\multicolumn{10}{@{}l}{\textsc{Prompt \& Multi-Agent-Based}} \\
\hspace*{1em}ReAct
  & 35.92\tabse{4.73} & 46.20\tabse{1.58} & 53.60\tabse{4.46} & 13.98\tabse{0.71} & 27.35\tabse{1.71} & 35.00\tabse{4.77} & 4.00\tabse{1.39} & 48.67\tabse{1.74} & 33.09\tabse{1.09} \\
\hspace*{1em}Least-to-Most
  & 37.86\tabse{4.78} & 47.40\tabse{1.58} & 48.00\tabse{4.47} & 14.40\tabse{0.71} & 27.21\tabse{1.71} & 39.00\tabse{4.88} & 5.00\tabse{1.54} & 51.70\tabse{1.74} & 33.82\tabse{1.10} \\
\hspace*{1em}ManuSearch
  & \secondbest{41.75}\tabse{4.86} & \secondbest{53.50}\tabse{1.58} & 60.80\tabse{4.37} & 16.84\tabse{0.76} & 28.38\tabse{1.73} & 45.00\tabse{4.97} & 7.50\tabse{1.86} & 62.14\tabse{1.69} & 39.49\tabse{1.12} \\
\midrule
\multicolumn{10}{@{}l}{\textsc{Single-Model Weight Adaptation}} \\
\hspace*{1em}Single LoRA
  & 35.92\tabse{4.73} & 49.70\tabse{1.58} & 54.40\tabse{4.45} & 15.35\tabse{0.73} & 35.15\tabse{1.83} & \secondbest{51.00}\tabse{5.00} & 12.50\tabse{2.34} & 55.83\tabse{1.73} & 38.73\tabse{1.13} \\
\hspace*{1em}Full SFT
  & \secondbest{41.75}\tabse{4.86} & 50.90\tabse{1.58} & \secondbest{65.60}\tabse{4.25} & 15.76\tabse{0.74} & \secondbest{38.53}\tabse{1.87} & 50.00\tabse{5.00} & 13.50\tabse{2.42} & \secondbest{66.38}\tabse{1.65} & 42.80\tabse{1.13} \\
\midrule
\multicolumn{10}{@{}l}{\textsc{Static Multi-LoRA Merging}} \\
\hspace*{1em}TIES-Merging
  & 34.95\tabse{4.70} & 47.60\tabse{1.58} & 55.20\tabse{4.45} & 14.07\tabse{0.71} & 34.41\tabse{1.82} & 44.00\tabse{4.96} & 7.00\tabse{1.80} & 48.06\tabse{1.74} & 35.66\tabse{1.11} \\
\hspace*{1em}DARE-TIES
  & 35.92\tabse{4.73} & 48.00\tabse{1.58} & 54.40\tabse{4.45} & 16.26\tabse{0.75} & 32.94\tabse{1.80} & 40.00\tabse{4.90} & 5.00\tabse{1.54} & 50.24\tabse{1.74} & 35.35\tabse{1.10} \\
\hspace*{1em}TSPA
  & 36.89\tabse{4.75} & 46.60\tabse{1.58} & 50.40\tabse{4.47} & 14.73\tabse{0.72} & 27.21\tabse{1.71} & 45.00\tabse{4.97} & 7.00\tabse{1.80} & 53.40\tabse{1.74} & 35.15\tabse{1.11} \\
\midrule
\multicolumn{10}{@{}l}{\textsc{\method{} (Ours)}} \\
\rowcolor{oursband}
\hspace*{1em}\method-hard
  & 40.78\tabse{4.84} & \best{56.30}\tabse{1.57} & \best{69.60}\tabse{4.11} & \secondbest{17.21}\tabse{0.77} & \best{40.00}\tabse{1.88} & \best{54.00}\tabse{4.98} & \best{17.50}\tabse{2.69} & \best{68.33}\tabse{1.62} & \best{45.47}\tabse{1.13} \\
\rowcolor{oursband}
\hspace*{1em}\method-soft
  & \best{46.60}\tabse{4.92} & 53.20\tabse{1.58} & 62.40\tabse{4.33} & \best{17.58}\tabse{0.77} & 37.79\tabse{1.86} & 50.00\tabse{5.00} & \secondbest{15.50}\tabse{2.56} & 62.62\tabse{1.69} & \secondbest{43.21}\tabse{1.14} \\
\bottomrule
\end{tabular}
\caption{Results on eight benchmarks (\%). Subscripts give nominal task-based binomial standard-error approximations. \textbf{Bold} and \underline{underlined} values denote the best and second-best scores per benchmark.}
\label{tab:main-results}
\end{table*}

The shared adapter matching the library's total parameter count has a mean absolute accuracy loss of 1.67\% across capabilities; extra capacity does not reduce this loss (Table~\ref{tab:capability-retention}). Writing gains least initially yet loses 49.2\% of that gain (Figure~\ref{fig:capability-retention}). Random partitioning still incurs a mean absolute accuracy loss of 1.12\%, so separation without capability alignment does not prevent forgetting.

\subsection{Capability Mismatch under Static Fusion}
\label{sec:exp:static-fusion}

We test whether static fusion preserves a fixed library of Plan, Search, and Writing specialists, each trained on its own capability. Full Additive sums their updates without final rank compression; TIES~\citep{tiesmerging}, DARE-TIES~\citep{dare,tiesmerging}, and TSPA~\citep{tspa} produce rank-64 updates. The reference, \emph{Individual}, selects the adapter matching each record's capability label. We reserve 200 records per capability as validation data and measure paired mean target-token negative log-likelihood (NLL) gaps under teacher forcing (fusion minus Individual). We also measure local mismatch through $(\Delta W_m-\Delta W_k)h$ on the same input activation $h$, where $\Delta W_m$ is the merged update and $\Delta W_k$ is the individual adapter update for capability $k$. For Full Additive, this difference is exactly the combined contribution of the other capability adapters. Validation data and metric details are in Appendix~\ref{app:merge-residual}.

None of the tested fusions eliminates the mean NLL degradation on any of the three capabilities (Figure~\ref{fig:merge-nll-distributions}). Local mismatch persists across LoRA layers in all three capabilities (Figure~\ref{fig:merge-layerwise}). Full Additive exhibits this mismatch despite retaining all updates without rank compression. Thus, combining intact updates can alter a specialist's local effect through contributions from other capabilities. These results support preserving the specialists as separate updates and selecting them at runtime.

\begin{figure*}[t]
\centering
\includegraphics[width=\textwidth]{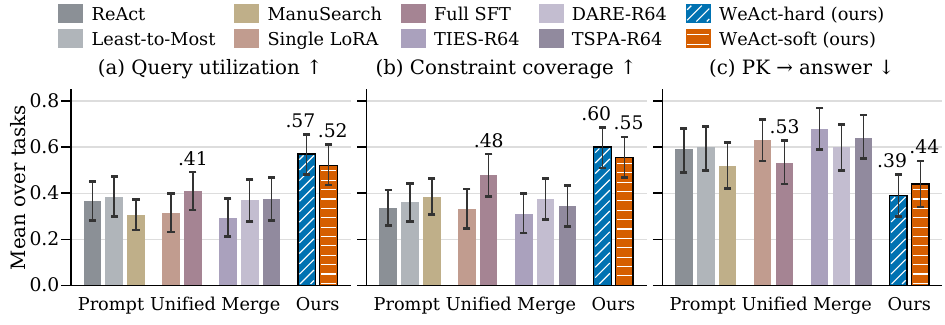}
\caption{SearchAtlas trace diagnostics. Query utilization: fraction of queries supporting the answer; constraint coverage: fraction of question constraints covered by those queries; PK$\rightarrow$answer: share of answers with unsupported content attributed to prior knowledge. Error bars are 95\% task-bootstrap intervals.}
\label{fig:trace-dag}
\end{figure*}

\endgroup

\section{Experiments}
\label{sec:experiments}

We evaluate \method on eight deep-research benchmarks, then analyze evidence use in execution traces and training and inference efficiency. We also test capability insertion and local updates to assess modular continual adaptation.

\subsection{Experimental Setup}
\label{sec:exp:setup}

\textbf{Model and adapters.} Weight-adaptation experiments start from \model{Qwen3-32B}~\citep{qwen3}: LoRA methods freeze the backbone; Full SFT updates it. Prompting baselines use the original checkpoint. The main \method library contains \plan{}, \search{}, and \writing{} with ranks 32, 64, and 32. Single LoRA uses rank 128, matching the combined parameter count of the three separately trained adapters; static merges use rank 64. Only capability expansion uses \code{}. Other training and deployment settings, including learning rates and epochs, are in Appendix~\ref{app:setup-details}.

\textbf{Training data.} We combine public agent trajectories from MiroVerse-v0.1~\citep{mirothinker2025} with supplementary trajectories generated from OpenSeeker-v1-Data~\citep{openseeker2026} questions using our execution framework. Each LoRA adapter's training samples retain execution context and supervise only its capability-specific outputs and handoff (Sec.~\ref{sec:training}). Appendix~\ref{app:segmentation} summarizes data sources and construction; continual adaptation studies describe additional Code and Writing data.

\textbf{Benchmarks.} We cover general tool use, multi-hop evidence composition, and web research. General uses \dataset{GAIA}~\citep{gaia}; Multi-Hop QA comprises \dataset{2WikiMultiHopQA}~\citep{2wikimultihopqa}, \dataset{Bamboogle}~\citep{press2023compositionality}, and \dataset{MuSiQue}~\citep{musique}. Search comprises \dataset{WebWalkerQA}~\citep{webwalkerqa}, \dataset{xbench-DeepSearch}~\citep{xbenchdeepsearch}, \dataset{BrowseComp}~\citep{browsecomp}, and \dataset{FRAMES}~\citep{frames}, covering web navigation, retrieval, hard-to-find facts, and multi-source integration.

\subsection{Main Results}
\label{sec:exp:main}

Table~\ref{tab:main-results} compares four adaptation mechanisms: prompting and multi-agent orchestration (ReAct~\citep{yao2023react}, Least-to-Most~\citep{leasttomost}, and ManuSearch~\citep{manusearch}); unified adaptation (Full SFT, Single LoRA); static fusion (TIES-Merging~\citep{tiesmerging}, DARE-TIES~\citep{dare,tiesmerging}, and TSPA~\citep{tspa}; Appendix~\ref{app:merge-residual}); and \method-hard/soft. Multi-agent methods keep their own context and model-call structures. Static fusion and \method share one capability library, fused into a fixed update or selected at runtime.

\method-hard has the highest average, 45.47\%, an accuracy gain of 2.67 points over Full SFT, and ranks first on six benchmarks. \method-soft has the second-highest average (43.21\%) and leads on GAIA and MuSiQue. The hard result supports independent LoRA learning with prescribed, stepwise invocation for reliable adapter use within one agent. Soft flexibly schedules independently trained LoRAs within the harness's runtime constraints.

Both variants outperform every static merge on all eight benchmarks; \method-hard also outperforms Single LoRA on all eight, and \method-soft on seven. Compared with Single LoRA, hard and soft achieve average-accuracy gains of 6.74\% and 4.48\%. The strongest static merge averages 35.66\%, below Single LoRA's 38.73\%: these rank-64 fusions of the independently trained adapters, whose ranks sum to 128, perform worse than direct rank-128 training. This gap is consistent with the fusion mismatch in Section~\ref{sec:exp:static-fusion}.

\subsection{Search Effectiveness}
\label{sec:execution-efficiency}
\label{sec:exp:part2}
\label{sec:exp:trace-analysis}

To assess search effectiveness beyond answer accuracy, we apply SearchAtlas~\citep{sang2026searchatlas} to all ten methods on the same 100 tasks, producing 1{,}000 evidential query graphs. \emph{Query utilization} ($\uparrow$) measures the fraction of issued queries on an evidence path to the answer; \emph{constraint coverage} ($\uparrow$), the fraction of question constraints represented in those queries; and \emph{PK$\rightarrow$answer} ($\downarrow$), the share of answers with content attributed to prior knowledge without retrieved support. Figure~\ref{fig:trace-dag} reports the results; Appendix~\ref{app:trace-analysis} gives formal definitions, graph construction, and task selection.

\method-hard ranks first and \method-soft second on all three metrics, ahead of Full SFT and the other baselines. Although ManuSearch issues the most queries, its answer-supporting queries cover only 39\% of question constraints, versus 60\% for \method-hard. These results show that \method connects search more effectively to answer support: a larger share of queries contributes to the answer, and those queries cover more question constraints. Fewer answers contain unsupported content attributed to prior knowledge.

\subsection{Training and Inference Efficiency}
\label{sec:exp:training-efficiency}
\label{sec:exp:rq2-efficiency}
\label{sec:exp:kv-ablation}

\textbf{Training compute.} We assess \method's training efficiency. With the same Qwen3-32B backbone and fewer epochs per adapter, \method uses 142.8 nominal EFLOPs, below Full SFT's training compute, while achieving higher average accuracy. We count repeated contexts, epochs, packing, dense operations, and causal attention, with 1.23$\times$ Full SFT's token exposure. LoRA omits frozen-backbone weight gradients. Across unequal capability segments, 2.06$\times$ token exposure at 0.95$\times$ per-token compute yields 1.96$\times$ total compute relative to Single LoRA with matched total trainable parameters (Appendix~\ref{app:rq2-training-profile}). These results support efficient training of independent capability adapters.

\begin{figure}[t]
\centering
\includegraphics[width=\columnwidth]{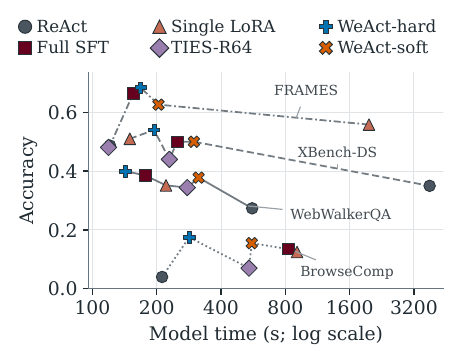}
\caption{Accuracy versus historical mean cumulative model-path time on four Search benchmarks. Times include saved retries.}

\label{fig:quality-model-time}
\end{figure}

\textbf{Inference efficiency.} Figure~\ref{fig:quality-model-time} pairs Table~\ref{tab:main-results} Search accuracy with historical mean cumulative model-path time on the same server with the same concurrency, harness, and tool conditions, including retries; General/Multi-hop QA appear in Figure~\ref{fig:quality-model-time-other} (Appendix~\ref{app:rq2-historical-efficiency}). Against Single LoRA, \method-soft is more accurate and faster on six benchmarks; hard improves accuracy on eight and time on seven. Across 50 runs per method on prespecified WebWalkerQA tasks with identical hardware and task concurrency one, soft and hard reduce mean answer-ready latency by 33.4\% and 20.3\% versus Full SFT; soft is within 1\% of Single LoRA. Model and tool execution dominate; adapter control and cache handling occupy 8.2\%/2.8\% of soft/hard time (Appendix~\ref{app:rq2-system}). These results support accuracy gains with modest switching overhead and net latency savings.

\suppressfloats[t]

\subsection{Modular Continual Adaptation}
\label{sec:continual-adaptation}
\label{sec:exp:rq3-maintenance}

We assess catastrophic forgetting and modular continual adaptation: expanding and refining the library while retaining existing capabilities. We first evaluate general-capability retention with external tools disabled (Appendix~\ref{app:preservation}), then examine Code insertion and a local Writing refresh.

\subsubsection{Continual Capability Expansion}
\label{sec:exp:code-extension}
\begin{table}[t]
\centering
\vspace{-0.4em}
\small
\setlength{\tabcolsep}{3.0pt}
\begin{tabular}{@{}lcccc@{}}
\toprule
Variant & MegaScience & SciBench & DS-1000 & $\Delta$GAIA \\
\midrule
w/o \texttt{LoRA-code} & 26.5 & 40.0 & 14.0 & 0.0 \\
w/ \texttt{LoRA-code} & \textbf{29.0} & \textbf{40.5} & \textbf{16.0} & \textbf{+1.0} \\
\bottomrule
\end{tabular}
\caption{Results of code-adapter insertion. $\Delta$GAIA is relative to the original $K=3$ system.}
\label{tab:rq3-code-extension}
\vspace{-0.8em}
\end{table}

To add code computation, we train a new \texttt{LoRA-code} adapter on MiroVerse-MegaScience trajectories converted to our tool and handoff protocol, keeping the base and existing adapters fixed without replay. Table~\ref{tab:rq3-code-extension} shows that inserting \texttt{LoRA-code} can improve accuracy on in-domain \dataset{MegaScience} and on \dataset{SciBench}~\citep{scibench}, without degrading \dataset{DS-1000}~\citep{ds1000} or \dataset{GAIA}~\citep{gaia}. Code insertion thus extends the library without retraining or harming existing capabilities (Appendix~\ref{app:code-extension}).

\subsubsection{Adapting to Evolving Writing Tasks}
\label{sec:exp:writing-refresh}

We test whether updating a single capability improves system-level report quality while preserving the rest of the library's parameters. We continue training the existing \writing{} adapter on trajectories synthesized in our harness from the OpenScholar~\citep{openscholar} subset of DR Tulu~\citep{drtulu} SFT data, replacing it with \writingplus{}. The base stays frozen, Plan and Search remain bitwise unchanged, and no historical data are replayed.

\begin{table}[t]
\centering
\fontsize{10}{12}\selectfont
\setlength{\tabcolsep}{2pt}
\begin{tabular*}{\columnwidth}{@{\extracolsep{\fill}}llcc@{}}
\toprule
\multicolumn{2}{c}{Method} & ResearchQA & \makecell[c]{DeepResearch\\Bench} \\
\midrule
\multirow{2}{*}{\writing{}} & hard & 58.33 & 26.51 \\
                           & soft & 54.14 & 27.92 \\
\addlinespace[2pt]
\multirow{2}{*}{\writingplus{}} & hard & 60.72 \textbf{(+2.40)} & 27.71 \textbf{(+1.21)} \\
                             & soft & 55.97 \textbf{(+1.83)} & 28.52 \textbf{(+0.61)} \\
\bottomrule
\end{tabular*}
\caption{Writing refresh scores. Bold values show gains over Original.}
\label{tab:writing-refresh-scores}
\end{table}

Table~\ref{tab:writing-refresh-scores} shows gains on \dataset{ResearchQA}~\citep{researchqa} and \dataset{DeepResearchBench}~\citep{deepresearchbench} under hard and soft routing: hard gains 2.40 and 1.21 points, respectively. Both variants improve all four DRB dimensions, with soft gaining most in readability and instruction following (Appendix~\ref{app:writing-refresh}, Table~\ref{tab:writing-refresh-rubrics}). Balancing new and historical Writing requirements within one module enables refinement without jointly retraining other capabilities. These gains support replay-free improvement without overwriting the backbone or other capability parameters.

\section{Related Work}
\label{sec:related-work}

\textbf{Deep-Research Agent Training.} WebSailor-V2 and Tongyi DeepResearch improve agent training through synthetic data and post-training~\citep{websailorv2,tongyideepresearch}. Step-DeepResearch and MiroThinker further organize learning around atomic capabilities~\citep{stepdeepresearch,mirothinker17}. These methods share parameters across capabilities, coupling subsequent updates. Separating trainable parameters allows local refinement without updating the remaining library; \method realizes this organization as callable adapters over a frozen backbone.

\textbf{Multi-Agent Orchestration.} ManuSearch delegates planning, search, and reading to specialized agents; Chain of Agents distributes long contexts across workers~\citep{manusearch,chainofagents}. Their handoffs require intermediate messages and a manager to integrate the workers' outputs. In \method, moving control between capability weights on a shared backbone offers another form of specialization: each selected policy continues from the same interaction history.

\textbf{Multi-LoRA Composition and Routing.} Early work such as LoRAHub and MoLE established reusable adapter composition~\citep{lorahub,lora-moe}. LD-MoLE learns token-dependent layerwise gates~\citep{ldmole}; LoGo selects and merges adapters per input without additional training~\citep{logo}. These methods focus on combining experts for task inputs rather than managing capability handoffs during an evolving agent trajectory. Adaptive Minds keeps the base model in control when calling adapters as tools~\citep{adaptiveminds2026}, whereas \method transfers control to the selected adapter. Macaron-V1 selects specialists per user turn~\citep{macaronv1}. Building on these approaches, \method couples independent capability learning with explicit handoffs within an agent trajectory. Selected adapters control reasoning and external tool use from shared history until offloading.

\section{Discussion}
\label{sec:discussion}

Research on preserving specialization spans the frozen columns of early Progressive Neural Networks~\citep{progressive2016} and the separate reasoning and tool-use LoRAs of recent DART~\citep{dart}. Our capacity-matched and random-partition controls show that capability alignment matters beyond capacity; fixed-library diagnostics reveal local fusion mismatch even without rank compression (Section~\ref{sec:capability-separation}). These results distinguish preserving learned parameters from preserving their functional effects during composition. Together with LoRA memory analysis~\citep{understandinglora2026}, they support aligning modules with independent update needs and selecting intact adapters as execution demands. Internal actions make this principle operational: newly trained LoRAs can enter an existing agent through the same interface. This suggests modular continual learning, with retention within updated modules remaining open.

\section{Conclusion}
\label{sec:conclusion}

We presented \method, which learns heterogeneous capabilities in independent LoRAs over a frozen backbone and switches active weights through internal actions within a shared agent trajectory. Controlled diagnostics support capability-aligned training and runtime selection for preserving specialization. Across eight deep-research benchmarks, \method improves average accuracy over Full SFT with lower nominal training compute. Code insertion expands the library with stable performance, and Writing refresh improves report quality. These results support independently learning agent capabilities and switching between them during execution, with potential for modular continual learning.

\section*{Limitations}
\label{sec:limitations}

Extending \method to sustained continual learning raises open questions about routing capacity and capability boundaries. It remains unclear how many adapters the agent can reliably distinguish and invoke as the library grows, and how well it can sustain longer sequences of internal actions. Repeated module insertion or refinement may also change capability selection, even when the rest of the library remains fixed. When task boundaries are unclear, general abilities such as reasoning may be needed across multiple stages, making it difficult to define separate LoRA modules and determine when each should be active. Whether incremental adapter expansion can improve these abilities while preserving performance on existing tasks remains an open question.

%\section*{Ethical Considerations}

%This work uses publicly available benchmarks and does not involve human-subject experiments, collection of personally identifying information, or deployment affecting individuals. We report the methods and limitations of the study transparently.

%The system generates reports and invokes external tools. In deployment, unsupported claims in reports or inappropriate tool calls could mislead users or expose sensitive information. Adapter switching alone does not verify evidence or enforce tool permissions. Deployment should therefore include access controls, privacy protections, and human verification of consequential outputs, consistent with the ACL Code of Ethics.\footnote{\url{https://www.aclweb.org/portal/content/acl-code-ethics}}

%\paragraph{AI assistance.}
% We used generative AI tools for language editing, code development, figure preparation, LaTeX formatting assistance, and bibliographic checks. We did not use generative AI tools to generate experimental results, fabricate data, or replace author verification of the methods and claims. All AI-assisted outputs were reviewed by the authors, who take responsibility for the final content of this work, including all text, claims, and artifacts.

\bibliography{custom}

\clearpage
\appendix
% Let appendix pages share wide floats while keeping the main-text policy.
\setcounter{dbltopnumber}{2}
% Use the same author-note visibility setting as the main text.
\section{Training Data Sources and Capability Segmentation}
\label{app:segmentation}

The training corpus contains 6,962 source trajectories before capability-specific splitting: 5,887 from MiroVerse-v0.1~\citep{mirothinker2025} and 1,075 generated from OpenSeeker-v1-Data~\citep{openseeker2026} questions using our execution framework (Table~\ref{tab:data-sources}).

\begin{table}[ht]
\centering
\fontsize{10}{12}\selectfont
\setlength{\tabcolsep}{3pt}
\begin{tabular}{@{}llr@{}}
\toprule
Use & Source & Count \\
\midrule
Main & MiroVerse-v0.1 & 5,887 \\
Main & OpenSeeker-v1-Data & 1,075 \\
\multicolumn{2}{@{}l}{\textbf{Main total}} & \textbf{6,962} \\
\midrule
Code & MiroVerse-MegaScience & 3,413 \\
Writing++ & DR-Tulu / OpenScholar & 1,000 \\
\bottomrule
\end{tabular}
\caption{Training-data sources. Main-corpus counts refer to source trajectories before capability splitting; Code and Writing++ counts refer to capability samples used in Section~\ref{sec:continual-adaptation}.}
\label{tab:data-sources}
\label{tab:span-stats}
\end{table}

\textbf{Full-SFT and capability training data.} Full SFT uses all 6,962 whole trajectories, covering all capabilities and retaining the five most recent tool responses. For \method, capability-specific splitting yields 6,962 Plan, 7,437 Search, and 6,962 Writing samples, with 200 samples per capability reserved for validation. Each sample retains the preceding execution context and supervises only the selected capability's assistant outputs and terminal offload call. The activation call and result belong to the masked prefix; all tool feedback is also masked from the loss.

Capability labels follow the stage's purpose (Section~\ref{sec:training}): Plan sets the goals and retrieval agenda; Search retrieves evidence, inspects sources, and cross-checks claims; Writing synthesizes the supported answer. A span may include several assistant/tool turns. The MiroVerse export separates planning before the first search from the subsequent search interaction and final answer synthesis. For continual adaptation, Code expansion extracts 3,413 code-computation samples from MiroVerse-MegaScience, with 200 reserved for validation; Writing refresh constructs 1,000 continuation samples from the OpenScholar subset of DR-Tulu SFT data~\citep{openscholar,drtulu}. These maintenance corpora are separate from the initial three adapters' data.

\textbf{Train--test overlap.} MiroVerse does not document the original splits of its sources, so we check overlap at the task level. We compare the task text of all 6,962 training trajectories with every evaluation question of the eight benchmarks. No evaluation question appears verbatim after normalization. A lexical screen flags questions with at least half of their IDF-weighted content words in a single training task whose trajectory also contains the gold answer; manual review of the flagged pairs finds overlap only through decomposed sub-agent subtasks. For 63 of the 2,417 MuSiQue questions (2.6\%), the answer-bearing sub-question appears with the same answer as a training subtask, and 19 of these restate the full question with the bridge entity resolved. One WebWalkerQA question and one 2Wiki question show similar overlap; GAIA, Bamboogle, XBench-DS, BrowseComp, and FRAMES show none. Removing the 63 MuSiQue questions lowers each method's MuSiQue accuracy by 0.04--0.47 points and leaves the ranking unchanged. The screen does not detect paraphrases below its threshold.

\subsection{A Search Training Example}
\label{app:training-example}

Figure~\ref{fig:training-example} shows a Search training sample for identifying which Republican candidates elected to the U.S. House of Representatives in 2008 previously served in the Minnesota House of Representatives. The boxed $x$ contains the system and user prompts, the Plan stage, and the Search loading exchange. It ends with the Search activation response, shown in full with headings and numbering added for readability. The boxed $y$ starts with the first Search assistant message and includes all five external exchanges and the final offload exchange. Its assistant tokens supply the targets in Eq.~\ref{eq:masked-sft}; $x$ and all tool feedback have zero loss. With \texttt{keep\_tool\_result=5}, the three tool returns in this prefix are all retained. Long contents and external-call arguments are abbreviated while all 20 messages remain visible.

% Standard ACL wide float; asymmetric panels accommodate the full activation response.
\begin{figure*}[p]
\centering
\begingroup
\colorlet{sampleInternal}{orange}
\definecolor{sampleExternal}{HTML}{246CA6}
\definecolor{samplemask}{HTML}{777777}
\colorlet{sampleInternalMasked}{sampleInternal}
\colorlet{sampleExternalMasked}{sampleExternal!65!white}
\lstset{language={},basicstyle=\fontfamily{cmtt}\fontsize{8.5}{9.3}\selectfont,
  columns=fullflexible,literate={_}{{\_\allowbreak}}1,
  keepspaces=true,showstringspaces=false,
  breaklines=true,breakatwhitespace=false,breakindent=8pt,
  frame=none,
  aboveskip=0pt,belowskip=0pt,xleftmargin=0pt,xrightmargin=0pt,
  postbreak={},upquote=true,
  moredelim={**[is][\color{sampleInternalMasked}]{|I|}{|/I|}},
  moredelim={**[is][\color{sampleExternalMasked}]{|E|}{|/E|}},
  moredelim={**[is][\fontfamily{pcr}\bfseries\selectfont]{|B|}{|/B|}}}
% Measure both listings once and give their outer frames the same height.
\setlength{\fboxsep}{5pt}
\setlength{\fboxrule}{0.5pt}
\newsavebox{\trainingXbox}
\newsavebox{\trainingYbox}
\newlength{\trainingBoxHeight}
\newlength{\trainingXWidth}
\newlength{\trainingYWidth}
\newlength{\trainingXInnerWidth}
\newlength{\trainingYInnerWidth}
\setlength{\trainingXWidth}{.66\textwidth}
\setlength{\trainingYWidth}{.31\textwidth}
\setlength{\trainingXInnerWidth}{\dimexpr\trainingXWidth-2\fboxsep-2\fboxrule\relax}
\setlength{\trainingYInnerWidth}{\dimexpr\trainingYWidth-2\fboxsep-2\fboxrule\relax}
\begin{lrbox}{\trainingXbox}
\begin{minipage}{\trainingXInnerWidth}
\begin{lstlisting}[basicstyle=\color{samplemask}\fontfamily{cmtt}\fontsize{8.5}{9.3}\selectfont]
[
  {"role": "system",
   "content": "[...] You may call exactly one function per
     assistant response [...] return a JSON object [...]
     within <tool_call></tool_call> XML tags:
     <tool_call>{\"name\": \"<function-name>\",
       \"arguments\": <args-json-object>}</tool_call>
     [...] Your goal is to provide complete, accurate and
     well-reasoned answers using additional tools. [...]"},
  {"role": "user",
   "content": "Check which of the Republican candidates elected
     to the U.S. House of Representatives in 2008 previously
     served in the Minnesota House of Representatives. [...]"},
  {"role": "assistant", "content": "[...] <tool_call>
     {\"name\": \"lora_runtime_mcp_server__load_lora\",
      \"arguments\": {\"lora_name\": \"lora-plan\"}}</tool_call>"},
  {"role": "tool", "content": "|I||B|[...]|/B||/I|"},
  {"role": "assistant", "content": "[...] <tool_call>
     {\"name\": \"lora_runtime_mcp_server__offload_lora\",
      \"arguments\": {\"lora_name\": \"lora-plan\"}}</tool_call>"},
  {"role": "tool", "content": "|I||B|[...]|/B||/I|"},
  {"role": "assistant", "content": "<think>The plan is ready.
     Now I need to gather evidence. I should activate the
     search stage.</think><tool_call>
     {\"name\": \"lora_runtime_mcp_server__load_lora\",
      \"arguments\": {\"lora_name\": \"lora-search\"}}</tool_call>"},
  {"role": "tool", "content": "|I|{
    \"success\": true,
    \"message\": \"LoRA session started: now uses 'lora-search'.
    # Stage activation confirmed: lora-search
    1. You are now responsible only for the current stage, not for
    the whole task from scratch.
    2. The base model has already selected this stage for the
    current step.
    3. Do not spend effort re-deciding global stage routing unless
    the runtime state or the user instruction changes.
    # Current stage objective:
    Close the current information gap by searching, browsing,
    reading, or scraping relevant external sources and extracting
    the facts needed for the next step.
    ## High-value outputs before offload:
    a relevant source, a retrieved fact set, or a short note
    that the current information gap is closed and the next stage
    can proceed
    ## Things usually not worth doing in this stage:
    1. restarting broad planning when the current step mainly
    needs evidence retrieval
    2. jumping straight to polished final prose while the key
    source or fact for this step is still missing
    ## Offload protocol
    1. When the current stage deliverable is complete, call
    `offload_lora` with `{\\\"lora_name\\\":
    \\\"lora-search\\\"}` and stop immediately.
    2. Do not combine that offload call with a final answer in the
    same response.
    3. When the current stage deliverable is complete, call:
    <tool_call>
    {\\\"name\\\": \\\"lora_runtime_mcp_server__offload_lora\\\",
    \\\"arguments\\\": {\\\"lora_name\\\": \\\"lora-search\\\"}}
    </tool_call>
    4. After that offload tool call, stop immediately.
    5. Do not combine the offload call with a final answer or a
    second tool call.\",
    \"active_lora\": \"lora-search\",
    \"no_op\": false
    }|/I|"}
]
\end{lstlisting}
\end{minipage}
\end{lrbox}
\begin{lrbox}{\trainingYbox}
\begin{minipage}{\trainingYInnerWidth}
\begin{lstlisting}[moredelim={**[is][\color{samplemask}]{|M|}{|/M|}}]
[
  {"role": "assistant",
   "content": "[...]
     <tool_call>
     {...}
     </tool_call>"},
  |M|{"role": "tool",
   "content": "|E||B|[...]|/B||/E|"},|/M|
  {"role": "assistant",
   "content": "[...]
     <tool_call>
     {...}
     </tool_call>"},
  |M|{"role": "tool",
   "content": "|E||B|[...]|/B||/E|"},|/M|
  {"role": "assistant",
   "content": "[...]
     <tool_call>
     {...}
     </tool_call>"},
  |M|{"role": "tool",
   "content": "|E||B|[...]|/B||/E|"},|/M|
  {"role": "assistant",
   "content": "[...]
     <tool_call>
     {...}
     </tool_call>"},
  |M|{"role": "tool",
   "content": "|E||B|[...]|/B||/E|"},|/M|
  {"role": "assistant",
   "content": "[...]
     <tool_call>
     {...}
     </tool_call>"},
  |M|{"role": "tool",
   "content": "|E||B|[...]|/B||/E|"},|/M|
  {"role": "assistant",
   "content": "<think>
     The search stage is
     complete. I will now
     offload the current
     LoRA and proceed to
     the next stage.
     </think>
     <tool_call>
     {
       \"name\":
         \"lora_runtime_mcp_server__offload_lora\",
       \"arguments\": {
         \"lora_name\":
           \"lora-search\"
       }
     }
     </tool_call>"},
  |M|{"role": "tool",
   "content": "|I|{
     \"success\": true,
     \"message\":
       \"LoRA session ended.
       Adapter unloaded
       from SGLang.\",
     \"active_lora\": null,
     \"server_unloaded\": true,
     \"no_op\": false
   }|/I|"}|/M|
]
\end{lstlisting}
\end{minipage}
\end{lrbox}
\setlength{\trainingBoxHeight}{\dimexpr\ht\trainingXbox+\dp\trainingXbox\relax}
\ifdim\dimexpr\ht\trainingYbox+\dp\trainingYbox\relax>\trainingBoxHeight
  \setlength{\trainingBoxHeight}{\dimexpr\ht\trainingYbox+\dp\trainingYbox\relax}
\fi
\begin{minipage}[t]{\trainingXWidth}
\raggedright\strut\textbf{$x$: retained context}\par\vspace{3pt}
\fcolorbox{black!50}{white}{\begin{minipage}[t][\trainingBoxHeight][t]{\trainingXInnerWidth}
\usebox{\trainingXbox}\vfill
\end{minipage}}
\end{minipage}\hfill
\begin{minipage}[t]{\trainingYWidth}
\raggedright\strut\textbf{$y$: Search continuation}\par\vspace{3pt}
\fcolorbox{black!50}{white}{\begin{minipage}[t][\trainingBoxHeight][t]{\trainingYInnerWidth}
\usebox{\trainingYbox}\vfill
\end{minipage}}
\end{minipage}
\endgroup
\caption{A training example for \texttt{lora-search}. Faded text is masked from the loss. Tool feedback is orange for internal actions and blue for external actions. $[\ldots]$ and $\{\ldots\}$ mark omitted text and arguments.}
\label{fig:training-example}
\end{figure*}

\subsection{Sensitivity to Capability Definitions and Annotation Rules}
\label{app:segmentation-sensitivity}

We explored capability definitions with GPT-5.4, starting from \texttt{Retrieval}, \texttt{Reasoning}, \texttt{Verification}, \texttt{Plan}, \texttt{Code}, and \texttt{Writing}. Paired comparisons use identical trajectories and assistant messages; changes count label-set differences, excluding boundary shifts within a label (Table~\ref{tab:segmentation-sensitivity}).

\begin{table}[t]
\centering
\small
\setlength{\tabcolsep}{3pt}
\begin{tabular}{@{}lrl@{}}
\toprule
\makecell[tl]{Policy change} & \makecell[tr]{Changed} & \makecell[tl]{Spans\\(before$\to$after)} \\
\midrule
\makecell[tl]{Strengthen targeted\\verification} & 4/80 & \makecell[tl]{Plan $20\to20$;\\Verif.\ $21\to21$} \\
\makecell[tl]{Narrow recap;\\broaden verification} & 36/78 & \makecell[tl]{Recap $30\to12$;\\Verif.\ $9\to41$} \\
\makecell[tl]{Revise tool/span\\rules} & 11/78 & \makecell[tl]{Evid.\ $19\to19$;\\Verif.\ $38\to37$} \\
\bottomrule
\end{tabular}
\caption{Policy sensitivity on fixed inputs. \emph{Changed}: relabeled/total assistant messages, including base-only transitions. Slices overlap and are not independent.}
\label{tab:segmentation-sensitivity}
\end{table}

\textbf{Fine-grained labels and competing definitions.} \emph{Separate \texttt{Code} from retrieval.} Prioritizing execution increased \texttt{Code} and reduced \texttt{Retrieval} assignments. \emph{Absorb retrieval into \texttt{Plan} or \texttt{Verification}.} Revised definitions shifted assignments between these labels, sometimes leaving aggregate counts unchanged. \emph{Isolate \texttt{State Recap}.} Recapping known facts and gaps overlapped with ongoing evidence gathering; narrowing recap and broadening verification changed many assignments. \emph{Replace recap with \texttt{Evidence Collection}.} This category remained ambiguous with \texttt{Verification}: identical lookups could be labeled as gathering details or checking a candidate.

\textbf{Granularity and category-specific rules.} Sparse labeling left nearly all messages unlabeled. Whole-turn labels recovered coverage; block-level labels also allowed opening plans and tool actions to be separated. Browse-oriented rules increased \texttt{Verification} on MiroVerse but reduced it on DR-Tulu while increasing \texttt{Plan}, indicating policy sensitivity rather than annotation accuracy. \emph{Introduce a \texttt{Math} capability.} \texttt{Math} received only one span in each larger MiroVerse and DR-Tulu batch, offering little support for a separate module in this corpus.

\textbf{Resulting capability design.} We retain \texttt{Plan}, \texttt{Search}, \texttt{Writing}, and \texttt{Code} for choosing goals and actions, acquiring and checking external evidence, synthesizing answers, and executing computations. \emph{Unify external-evidence operations as \texttt{Search}.} Evidence collection, source inspection, and verification share one capability, avoiding adapter assignments based on how a lookup is phrased. The main-table library uses \texttt{Plan}, \texttt{Search}, and \texttt{Writing}; \texttt{Code} is tested separately through capability insertion. The initial merged \texttt{Search} variant covered fewer tool-call blocks than its predecessor. Structural search candidates and text-anchored boundaries were still needed to recover coverage. These annotation findings motivate the chosen capabilities and segmentation procedure.

\section{Training and Inference Efficiency Details}
\label{app:rq2-efficiency}

\subsection{Nominal Training Compute}
\label{app:rq2-training-profile}

All recipes use the same Qwen3-32B backbone and a maximum sequence length of 32k. Single LoRA matches the three adapters' total trainable parameters: rank 128 versus ranks 32/64/32 on the same target projections. Plan, Search, and Writing train for two, three, and three epochs, respectively; Single LoRA trains for three and Full SFT for five. We account for each recipe's actual token exposure and sequence lengths.

We count major matrix operations in Qwen3-32B training, with two FLOPs per multiply-add. Let $D_m$ count processed token positions and $Q_m$ estimate the sum of squared unpadded sequence lengths for recipe $m$. With dense weights $P$, rank-$r$ adapter weights $P_r$, $L$ layers, and total query width $a$,
\begin{equation}
\begin{gathered}
F_{\mathrm{Full}}=6PD_{\mathrm{Full}}+6LaQ_{\mathrm{Full}},\\
F_m=(4P+6P_{r_m})D_m+6LaQ_m,\\
F_{\mathrm{WeAct}}=\!\sum_{m\in\{\mathrm{plan,search,writing}\}}F_m,\\
P=31.984\times10^9,\quad P_r=8.546\times10^6r,\\
L=64,\quad a=8192.
\end{gathered}
\label{eq:training-compute}
\end{equation}
In $F_{\mathrm{Full}}$, $6PD_{\mathrm{Full}}$ counts forward, input-gradient, and weight-gradient products; in $F_m$, $4PD_m$ omits frozen weight gradients; $6P_{r_m}D_m$ trains the adapter; the shared $6LaQ_m$ term counts causal query--key and attention--value products in both passes; $F_{\mathrm{WeAct}}$ sums the three independently trained capability adapters' costs; $P$ sums dense matrix sizes; $P_r$ sums rank-$r$ factor sizes, including the output head; $L$ and $a$ are Qwen3-32B's layer count and total query width.

\begin{figure*}[t]
\centering
\includegraphics[width=0.9\linewidth]{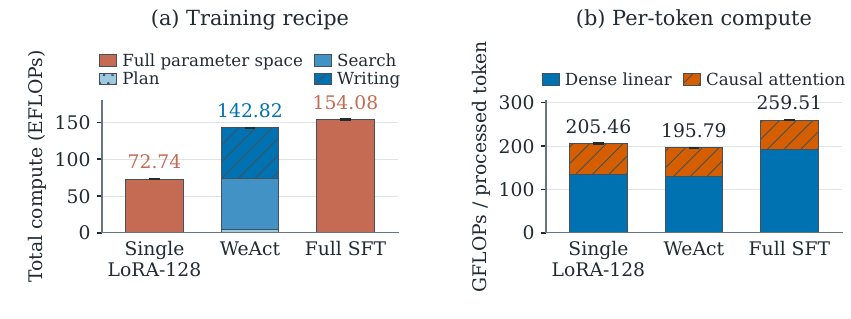}
\caption{Nominal training compute. (a) Total recipe compute, with \method split by capability. (b) Dense and causal-attention FLOPs per token. Error bars show length-statistic uncertainty (details in text).}
\label{fig:training-compute}
\end{figure*}

Token exposure includes repeated contexts and epochs. Packed padding contributes to dense operations; block-diagonal attention uses individual unpadded trajectories. Length statistics retain the training text format, including reasoning; Single LoRA's squared-length sum is estimated from randomized batch-token variance. Tool-call learning is part of adapter training; no separate router is trained. We exclude elementwise operations, optimizer updates, activation recomputation, communication, and I/O. Figure~\ref{fig:training-compute} shows length-statistic uncertainty as 95\% bootstrap intervals for Single LoRA and Full SFT and summed component sensitivity bounds for \method.

\method uses 142.8 EFLOPs, 7.3\% below Full SFT, despite processing 1.23$\times$ as many tokens (Figure~\ref{fig:training-compute}). Relative to Single LoRA, 2.06$\times$ token exposure at 0.95$\times$ the per-token cost gives 1.96$\times$ total compute. Independent capability training thus costs less than Full SFT under these recipes, while its greater token exposure raises total compute over Single LoRA. These estimates use each recipe's actual sequence lengths and padding, rather than measuring hardware throughput.

\subsection{Historical Inference Efficiency}
\label{app:rq2-historical-efficiency}

\begin{figure*}[t]
\centering
\includegraphics[width=13cm]{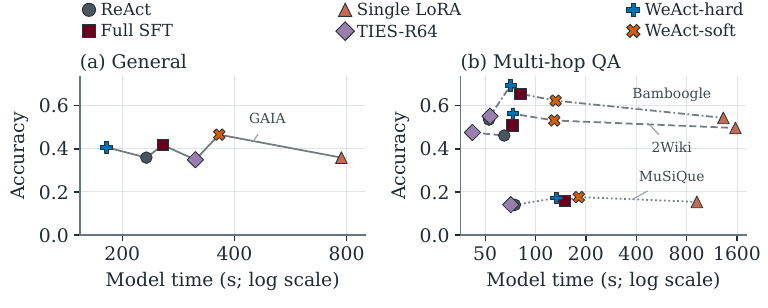}
\caption{Accuracy versus model-path time on General and Multi-hop QA. Points pair Table~\ref{tab:main-results} accuracy with historical mean cumulative time, including retries. Colors and markers identify methods; line styles group benchmarks. Time axes are logarithmic in seconds.}
\label{fig:quality-model-time-other}
\end{figure*}

\textbf{Records and timing.} This analysis uses the logged runs behind the main results, collected over time on the same server with the same concurrency, harness, and tool conditions; we call them historical runs. The quality--time analysis covers six methods on eight benchmarks. The wall-time ECDFs additionally include ManuSearch, whose BrowseComp logs are unavailable, and use all 37,928 task-level records with valid timing. For each task and attempt, the first run is retry 0 and the final run is the highest retry index. Accuracy uses the final run's judgment, and cost sums all runs up to it. Figures~\ref{fig:quality-model-time} and~\ref{fig:quality-model-time-other} pair these times with the accuracy in Table~\ref{tab:main-results}.

Model-path time is the union of model-call intervals within a run, summed across retries. Task-record wall time spans the logged task start and end and can include post-answer judging, so it measures logged execution time rather than answer-ready latency. The ECDFs include successful, failed, and anomaly-flagged records with valid timing, with one curve per method and benchmark.

\textbf{Quality--model-path time comparisons.} On GAIA, \method-soft improves accuracy over Single LoRA from 35.9\% to 46.6\% while reducing mean model-path time from 774.9 to 364.1 seconds; \method-hard reaches 40.8\% at 181.0 seconds (Figure~\ref{fig:quality-model-time-other}). TIES-R64 has shorter model paths on Multi-hop QA but lower accuracy than both \method variants on all eight benchmarks.

\begin{figure*}[t]
\centering
\includegraphics[width=\linewidth]{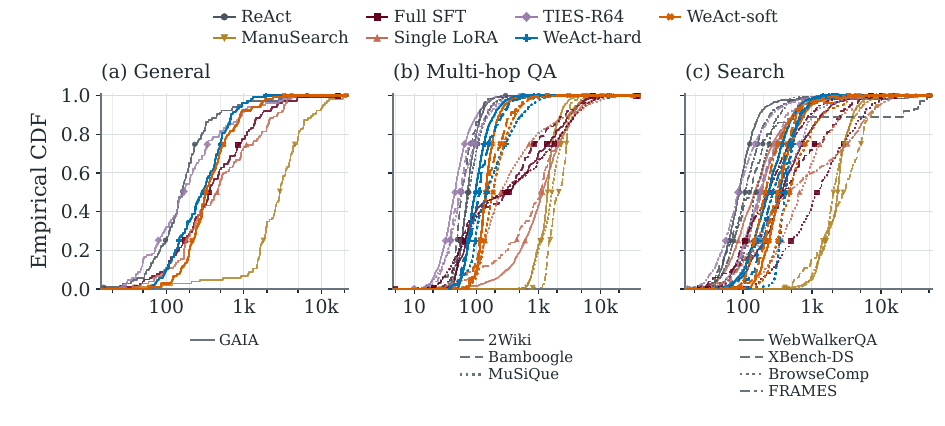}
\caption{Historical task-record wall-clock distributions. Colors and markers identify methods; line styles identify benchmarks. Each panel uses its own logarithmic time axis in seconds and retains all tails. Curves include all valid task-record times and saved retries, regardless of correctness; logged wall time can include post-answer judging.}
\label{fig:historical-wallclock}
\end{figure*}

\textbf{Typical cost and long tails.} Both \method variants have lower mean task-record wall time than Full SFT on all eight benchmarks and lower means than Single LoRA on six and seven benchmarks for soft and hard, respectively. ReAct and TIES-R64 often have shorter typical runs, but \method has much shorter tails (Figure~\ref{fig:historical-wallclock}). On GAIA, their medians are lower than both \method variants', yet \method-hard's 99th percentile is below those of all four baselines.

On Multi-hop QA, ReAct and TIES-R64 also have shorter typical runs, while \method is more accurate than both and uses less mean wall time than Full SFT and Single LoRA. The largest tails occur on Search: on WebWalkerQA, \method-hard's 99th percentile is 1,092 seconds versus ReAct's 25,315 seconds, despite a longer median. XBench-DS shows the same contrast, although Single LoRA has a lower mean than \method-hard at lower accuracy. On the seven benchmarks with ManuSearch logs, both \method variants have lower mean and median wall time than ManuSearch. Thus the mean-time gains coexist with longer typical runs than some baselines and substantially shorter extreme runs.

\textbf{Retries and first-run performance.} Table~\ref{tab:historical-retries} reports the share of tasks with saved retries on GAIA, 2Wiki, and WebWalkerQA. \method-soft retries more tasks than \method-hard, particularly on WebWalkerQA. \method-hard retries fewer tasks than the three trained baselines, and its accuracy changes by at most 0.15 percentage points from first to final run. By comparison, Single LoRA's GAIA accuracy rises from 28.2\% to 35.9\%.

\begin{table*}[tp]
\centering
\fontsize{10}{12}\selectfont
\setlength{\tabcolsep}{4pt}

\begin{tabular}{lrrrrrrr}
\toprule
Benchmark & $N$ & ReAct & Full SFT & Single LoRA & TIES-R64 & \method-soft & \method-hard \\
\midrule
GAIA & 103 & 0.00 & 16.50 & 26.21 & 14.56 & 4.85 & 1.94 \\
2Wiki & 1,000 & 0.70 & 1.00 & 6.40 & 4.70 & 1.00 & 0.60 \\
WebWalkerQA & 680 & 0.29 & 9.26 & 11.32 & 12.79 & 9.26 & 3.82 \\
\bottomrule
\end{tabular}
\caption{Tasks with saved retries (\%). The denominator is the task count $N$ for each method and benchmark.}
\label{tab:historical-retries}
\end{table*}

For \method-hard, the model-path time advantage over Single LoRA is already present in first runs on GAIA and 2Wiki. On WebWalkerQA, \method-hard uses 0.94$\times$ Single LoRA's first-run time, compared with 0.65$\times$ when retries are included, so most of that gap comes from retries. The logs do not identify the cause of each retry.

\subsection{Runtime Decomposition and Cache Costs}
\label{app:rq2-system}
\label{app:rq2-system-cost}

We evaluate Full SFT, Single LoRA, \method-hard, and \method-soft on the same prespecified WebWalkerQA tasks, with 50 runs per method on the same hardware at task concurrency one. All runs are included regardless of correctness. Answer-ready time spans the complete root-agent execution, excluding subsequent judging. We partition this interval into mutually exclusive model, external-tool, adapter-control, observed cache, answer-extraction, and setup/orchestration intervals; their sum equals the root duration for every run. Model time is further separated by the active adapter or base route. Identified cache intervals, covering the observed cache flush and post-switch prefill, are moved out of their enclosing model calls so that no time is counted twice.

\begin{figure*}[t]
\centering
\includegraphics[width=\linewidth]{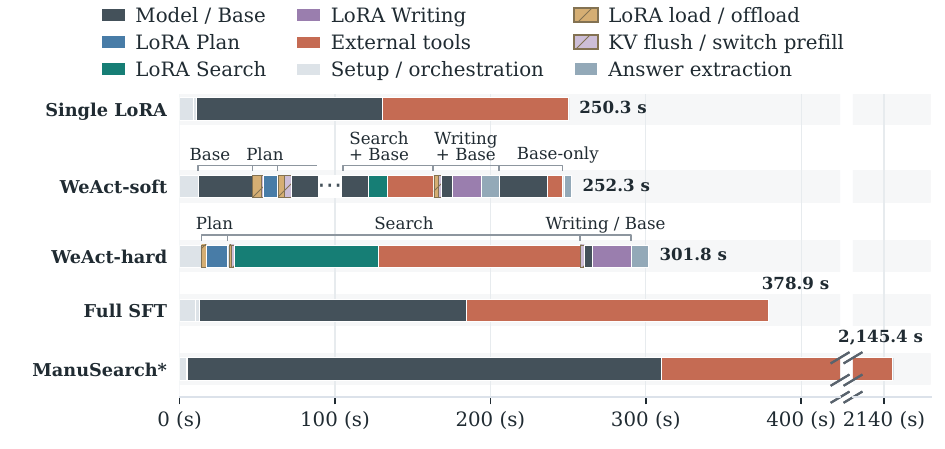}
\caption{Runtime decomposition on WebWalkerQA with 50 runs per method. Bars show stage-aligned means. The ellipsis denotes repeated capability switching; all durations are retained. ManuSearch* is a historical reference with unmatched deployment and tool conditions.}
\label{fig:runtime-decomposition}
\end{figure*}

Colors show the mean composition within each stage, not a single run's event order. For \method-hard, the stages are Plan, Search, and a final Writing or Base stage. For \method-soft, Plan spans its load--offload interval, Search runs from the Plan offload to the Writing load (or to the end when Writing is absent), and Writing covers the remainder; these stages include Base calls and tools and average 16.2, 100.0, and 42.5 seconds, respectively. Terminal extraction, the startup Base interval, and Base-only execution are shown separately.

Figure~\ref{fig:runtime-decomposition} shows that \method-soft's mean latency is close to Single LoRA's, while both \method variants use less time than Full SFT. Adapter control and observed cache intervals together occupy 8.2\% of soft's time and 2.8\% of hard's. Time is attributed to the model that served each request, so base-model and external-tool time within a Search stage is not counted as Search-adapter time.

For the ManuSearch reference, we use 50 first-attempt runs without retries on the same task set, after checking that the question text matches. External tools account for 85.6\% of its mean answer-ready time and model calls for 14.2\%; extraction or summarization performed inside a tool counts as tool time. The figure breaks its tool bar between 425 and 2,120 seconds, using the same scale on both sides.

\section{Sequential Capability Retention}
\label{app:sequential-retention}

\textbf{Adapter organizations.} All four organizations start from frozen \model{Qwen3-32B} with identical initial weights. The shared baselines train one LoRA of rank 32 or 96, and the partitioned libraries train three rank-32 adapters. The capability partition gives each capability its own adapter in every trajectory. The random partition draws a fixed capability-to-adapter permutation for each trajectory, balanced across trajectories and kept for training and evaluation, so every adapter receives data from all three capabilities. A uniform rank isolates how parameters are organized; the main library uses capability-specific ranks (Section~\ref{sec:exp:setup}).

\textbf{Data.} We use 384 training and 192 validation trajectories, partitioning the latter into 168 for stage-end evaluation and 24 for monitoring. For each trajectory and capability, a fixed hash selects one complete assistant message as the target. The message is split into chunks of at most 512 target tokens, each with up to 7,168 preceding tokens of context; longer contexts keep the initial and most recent tokens around an omission marker. Each chunk with its context forms one training window.

\textbf{Optimization.} LoRA is applied to all attention and feed-forward projections with $\alpha=r$ and no dropout. Trainable parameters and optimizer moments are kept in FP32 over the BF16 backbone. Each order runs 288 updates, 96 per stage, with six windows per update; all organizations and orders consume the same window streams. We use AdamW without weight decay or gradient clipping, a linear warmup over the first 16 updates to $2\times10^{-4}$, and a constant rate afterwards. Optimizer state persists across stages, and adapters inactive in a stage receive no updates. All 24 runs (four organizations by six orders) use one training seed, so the six orders are experimental conditions rather than replicates.

\begin{figure*}[t]
\centering
\includegraphics[width=\linewidth]{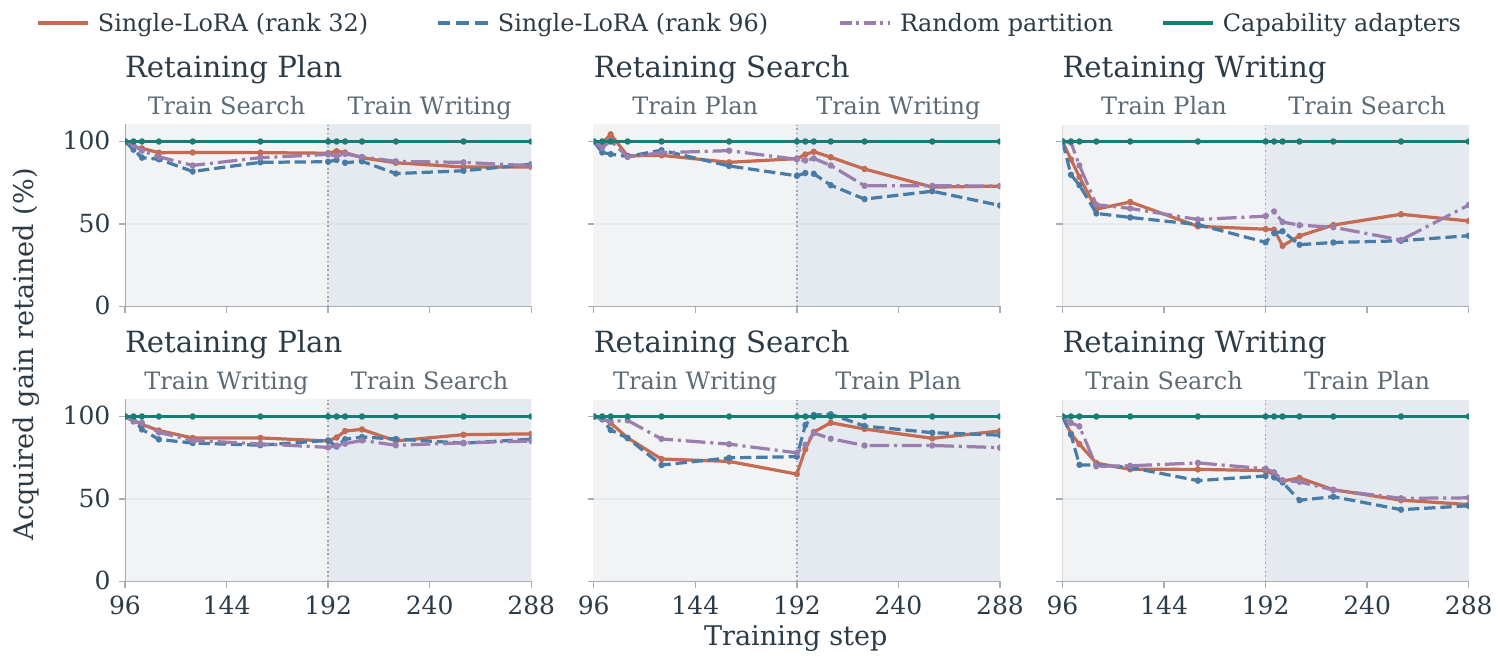}
\caption{Retention during subsequent training. Panels track the first capability in each order on monitoring trajectories; shading marks later stages, and 100\% denotes the gain acquired in its own stage.}
\label{fig:capability-retention-dynamics}
\end{figure*}

\textbf{Evaluation.} Content-token accuracy is teacher-forced top-1 accuracy on the target tokens, excluding special tokens. We pool tokens within each message and weight trajectories equally. We evaluate the 168-trajectory validation subset before training and at the end of each stage (steps 0, 96, 192, and 288); the separate 24-trajectory subset is used only for the monitoring curves in Figure~\ref{fig:capability-retention-dynamics}. Let $A_k^0$, $A_k^{\mathrm{own}}$, and $A_k^{\mathrm{final}}$ be the validation accuracy of capability $k$ before training, at the end of its own stage, and after all stages. Averaging each over validation trajectories and over the four orders in which $k$ precedes another capability, we compute the fraction of acquired gain lost, $F_k$:
\begin{equation}
F_k
  = \frac{A_k^{\mathrm{own}}-A_k^{\mathrm{final}}}
         {A_k^{\mathrm{own}}-A_k^0}.
\label{eq:acquired-gain-lost}
\end{equation}
The ratio is computed from averaged accuracies rather than averaged over trajectories. Table~\ref{tab:capability-retention} reports its denominator (acquired gain), numerator (lost gain), and value.

\begin{table*}[tp]
\centering
\fontsize{10}{12}\selectfont
\setlength{\tabcolsep}{3pt}

\begin{tabular}{lrrrrrrrrr}
\toprule
 & \multicolumn{3}{c}{Plan} & \multicolumn{3}{c}{Search} & \multicolumn{3}{c}{Writing} \\
\cmidrule(lr){2-4}\cmidrule(lr){5-7}\cmidrule(lr){8-10}
Adapter organization & G & L & \% & G & L & \% & G & L & \% \\
\midrule
Single-LoRA (rank 32) & 14.28 & 1.67 & 11.7 & 6.95 & 1.57 & 22.6 & 3.05 & 1.41 & 46.2 \\
Single-LoRA (rank 96) & 14.31 & 1.82 & 12.7 & 6.81 & 1.68 & 24.7 & 3.06 & 1.50 & 49.2 \\
Random partition & 13.36 & 1.32 & 9.9 & 6.16 & 0.96 & 15.6 & 2.58 & 1.09 & 42.4 \\
Capability adapters & 14.22 & 0.00 & 0.0 & 6.84 & 0.00 & 0.0 & 3.09 & 0.00 & 0.0 \\
\bottomrule
\end{tabular}
\caption{Acquired gain (G) and lost gain (L) in points of content-token accuracy, and fraction of acquired gain lost (\%), pooled over the four orders in which each capability precedes another. Fractions use unrounded means.}
\label{tab:capability-retention}
\end{table*}

\textbf{Retention during later stages.} Figure~\ref{fig:capability-retention-dynamics} follows the first capability of each order on the monitoring trajectories. Writing falls to about 55--70\% of its gain within the first few checkpoints of the next stage, whereas Plan keeps roughly 80--95\% throughout. Search loses more while Writing is trained than while Plan is trained, and partly recovers when Plan training follows Writing. The shared and randomly partitioned organizations follow similar trajectories.

\section{Diagnostic Metrics for Capability Separation}

\label{app:rq1-metrics}
% not checked yet
\begin{figure*}[t]
\centering
\includegraphics[width=0.84\linewidth]{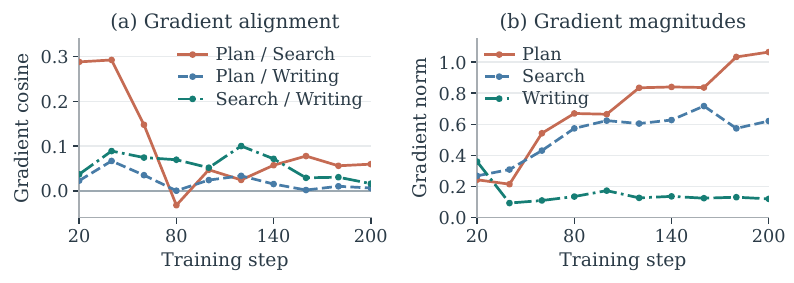}
\caption{Gradient alignment and magnitude in the Single-LoRA pilot. (a)~Pairwise cosines and (b)~gradient norms at ten checkpoints of a rank-96 Single-LoRA run on frozen \model{Qwen3-32B}. Most pairwise cosines are positive, while gradient magnitudes differ across capabilities.}
\label{fig:rq1-diagnostic}
\end{figure*}

\textbf{First-order update analysis.} Eq.~\ref{eq:shared-update} considers a step on a single capability. For joint training, let $\mathcal D_1,\dots,\mathcal D_K$ be the span datasets with sampling shares $p_k$ and per-capability losses $\mathcal L_k^{\mathrm{shared}}(\omega)$ over shared parameters $\omega$. A shared adapter minimizes $\mathcal L_{\mathrm{mix}}^{\mathrm{shared}}(\omega) =\sum_k p_k\mathcal L_k^{\mathrm{shared}}(\omega)$ with gradient steps $\delta\omega=-\eta\sum_k p_k g_k$, where $g_k=\nabla_\omega\mathcal L_k^{\mathrm{shared}}(\omega)$. Writing $\Delta\mathcal L_i$ for the change in capability $i$'s shared-adapter loss, the first-order approximation is
\begin{equation}
\Delta\mathcal L_i \approx -\eta\Bigl(p_i\norm{g_i}^2
  + \textstyle\sum_{j\neq i}p_j\inner{g_i}{g_j}\Bigr).
\label{eq:mixture-decomp}
\end{equation}
When $\inner{g_i}{g_j}\approx0$, the other capabilities' updates give capability $i$ little first-order transfer, and its own term scales with $p_i$, so unequal sampling shares or gradient norms let one capability dominate the mixed direction. Like Eq.~\ref{eq:shared-update}, this describes one gradient step; with an adaptive optimizer, the realized update also depends on optimizer state.

\textbf{Gradient measurements.} For capability gradients $g_i$ and $g_j$ in the same adapter parameter space, we report $\cos(g_i,g_j)=\inner{g_i}{g_j}/(\norm{g_i}\norm{g_j})$ and the norms $\norm{g_i}$. The norm ratio $\max_i\norm{g_i}/\min_i\norm{g_i}$ is computed without weighting by sampling share, which enters through Eq.~\ref{eq:mixture-decomp}.

Figure~\ref{fig:rq1-diagnostic} probes a jointly trained rank-96 Single LoRA ($\alpha=32$) at ten evenly spaced training checkpoints, using the same pool of capability spans at every checkpoint. Most cosines remain positive, while the largest-to-smallest norm ratio grows over training; Writing has the smallest gradients after the first checkpoint. Positive cosines mean that, to first order, a step on one capability does not raise another's loss. The loss in Section~\ref{sec:exp:sequential-retention} arises under sequential training, where a capability that has already been fit has small gradients, so later updates can raise its loss through higher-order terms even without gradient conflict.

\section{Static-Merge Residual Details}
\label{app:merge-residual}

\subsection{Fixed Capability Library and Validation Data}

We fix \model{Qwen3-32B} and the same independently trained Plan, Search, and Writing adapters throughout this diagnostic. Adapter ranks and scaling follow the main experimental setup (Appendix~\ref{app:setup-details}). Individual activates the adapter matching the target capability. Full Additive sums the effective updates $\Delta W_k=(\alpha_k/r_k)B_kA_k$ without final rank compression. TIES-R64, DARE-TIES-R64, and TSPA-R64 construct a fixed merged update and represent it at rank 64. TIES uses density 0.2; DARE-TIES additionally uses drop rate 0.5 and seed 0. We follow the original TSPA procedure~\citep{tspa}, which performs attention alignment followed by LoRA parameter alignment. It optimizes rotations and scaling within the low-rank factors before merging the aligned adapters.

We use our capability-specific validation split: each capability's source data is shuffled with seed 42, the first 200 records form each capability's validation set, and the remainder is used for training. Both the activation and NLL diagnostics draw their evaluation records from these validation sets. We render their messages with the training format, preserving supervision masks, capability labels, and recorded tool observations. Evaluation replays the recorded text without making new external tool calls. The context and target selection rules for each diagnostic are described below.

\subsection{Local Effects on Matched Activation}

The activation diagnostic tests whether a static update preserves the local effect of the capability-specific update. For projection $j$, we run Individual to obtain its input $h_{k,j}$ and evaluate every candidate on that same input:
\begin{equation}
\rho_{m,k,j}=
\frac{\| (\Delta W_{m,j}-\Delta W_{k,j})h_{k,j}\|_2}
{\max\!\left(\|\Delta W_{k,j}h_{k,j}\|_2,10^{-12}\right)}.
\label{eq:activation-conditioned-residual}
\end{equation}
Here $m$ identifies the merge rule, $k$ the target capability, and $j$ the LoRA projection; $\Delta W_{m,j}$ and $\Delta W_{k,j}$ are the merged and individual adapter updates, respectively. This measures adapter-induced local mismatch, excluding differences introduced by propagating each candidate's own activations through later layers. For Full Additive, writing $h=h_{k,j}$, the difference is exactly the combined contribution of the other capability adapters:
\begin{equation}
\textstyle\bigl(\sum_i\Delta W_{i,j}-\Delta W_{k,j}\bigr)h
=\sum_{i\ne k}\Delta W_{i,j}h.
\label{eq:merge-nontarget-contribution}
\end{equation}
The residual thus measures how non-target updates change the matched adapter's local action; their contributions can partly reinforce or cancel one another.

We retain an 8,192-token prefix of each trajectory and select causally aligned supervised positions. The activation used to predict a target token is taken from the preceding position, with its capability label checked at that source position. Each validation record contributes four positions across all attention and feed-forward LoRA projections in every transformer layer. We average positions and projections within each layer and trajectory, then weight trajectories equally.

\begin{figure*}[t]
\centering
\includegraphics[width=\linewidth]{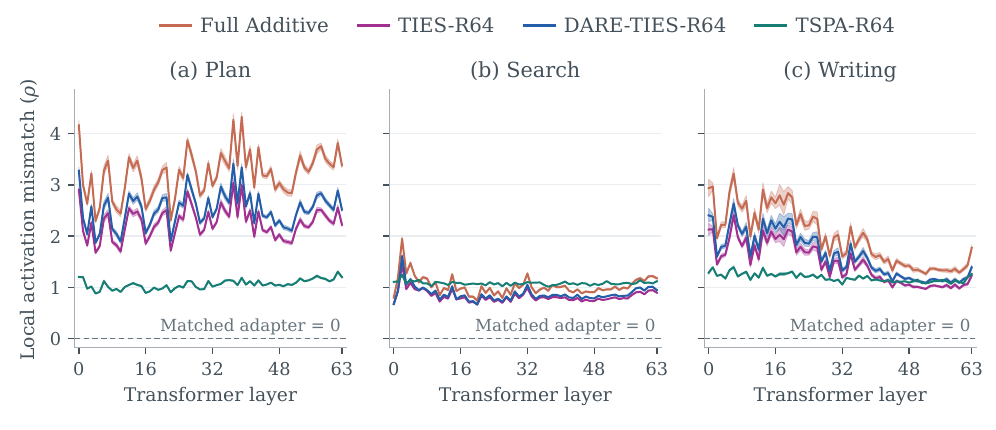}
\caption{Local mismatch on capability-matched activations. Curves show trajectory-weighted layer means of Eq.~\ref{eq:activation-conditioned-residual}; shading gives pointwise 95\% trajectory-bootstrap confidence intervals. The dashed zero line is Individual's identity reference. Full Additive uses no final rank compression.}
\label{fig:merge-layerwise}
\end{figure*}

Figure~\ref{fig:merge-layerwise} shows capability-dependent local mismatch across layers. Conflict handling and subspace alignment change its magnitude, and the ordering of the fusions varies by capability, but none recovers the matched update's local effect for all three capabilities.

\subsection{Target-Text Fit and Variation across Trajectories}

\begin{figure*}[t]
\centering
\includegraphics[width=0.95\linewidth]{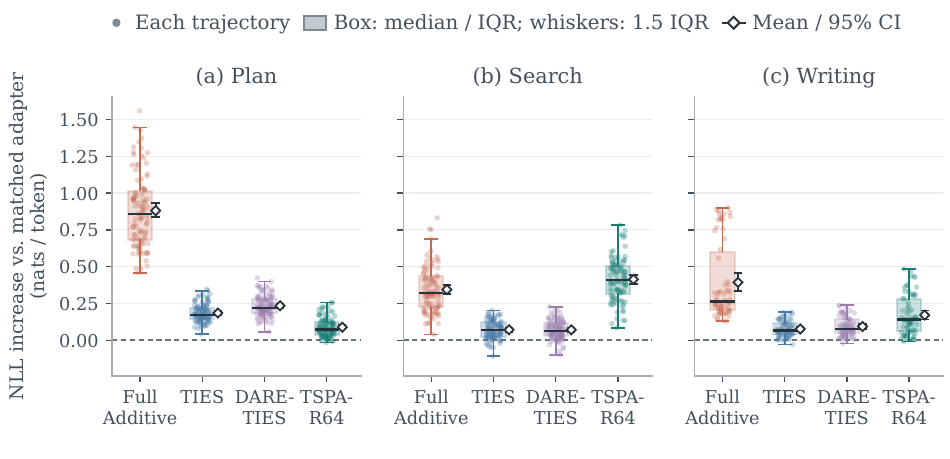}
\caption{Paired target-token NLL gaps relative to Individual. Points show validation records, boxes show medians and interquartile ranges, and diamonds mark means with 95\% confidence intervals.}
\label{fig:merge-nll-distributions}
\end{figure*}

We complement the local diagnostic with teacher-forced target-token NLL, comparing each static condition and Individual on the same record and supervised targets. LoRA updates are applied to all adapted layers, including the output head. Contexts are capped at 4,096 tokens, and each record contributes the mean NLL of the valid supervised tokens visible in its target capability span. For record $x$ with capability $k$ and merge rule $m$, we compute
\begin{equation}
\Delta L_{m,k}(x)=L_m(x)-L_k(x),
\label{eq:merge-nll-gap}
\end{equation}
where $L$ is the record's mean target-token NLL. Figure~\ref{fig:merge-nll-distributions} shows the distribution of these paired differences across validation records, with diamonds marking their equally weighted means; zero denotes equal target-text fit.

The mean NLL gap is positive for every tested fusion in all three capabilities, although some records have gaps close to zero or negative (Figure~\ref{fig:merge-nll-distributions}). Full Additive's Writing mean exceeds its median, indicating that a subset of records with large gaps raises the mean. The tested fusions therefore reduce mean target-text fit, with the size and direction of the change varying across records.

The local and NLL diagnostics agree at the capability level: every tested fusion leaves local mismatch and poorer mean target-text fit relative to capability-matched use. These results support preserving the specialists as separate updates and selecting them at runtime. Section~\ref{sec:exp:main} reports the task-level effect of these merges.

\section{Experimental Setup Details}
\label{app:setup-details}

This appendix gives the configuration of the main experiments; the sequential diagnostic uses its own configuration (Appendix~\ref{app:sequential-retention}).

\textbf{Capability adapters.} The capability adapters use ranks 32, 64, and 32 for Plan, Search, and Writing, with $\alpha_k=32$ and no LoRA dropout. Search uses rank 64 because rank-32 pilots showed repetitive, non-terminating reasoning and unreliable external-tool and \texttt{offload\_lora} calls. Each adapter places low-rank factors on the query, key, value, and output projections and on the gate, up, and down projections of all 64 transformer layers, and on the output projection (\texttt{lm\_head}); input embeddings are not adapted. We train the adapters with Tinker~\citep{tinker} using the Qwen3 chat template, a batch size of eight, and a maximum sequence length of 32k. Adam uses $(\beta_1,\beta_2)=(0.9,0.95)$ and $\epsilon=10^{-8}$, and the learning rate decays from its peak to zero on a cosine schedule without warmup. Plan trains for two epochs at a peak learning rate of $2\times10^{-4}$; Search and Writing train for three epochs at $3\times10^{-4}$ and $2\times10^{-4}$, respectively. Each capability's data are shuffled with seed 42, and the first 200 records form a validation set for NLL monitoring. The activation and NLL diagnostics draw their evaluation records from these validation sets (Appendix~\ref{app:merge-residual}). Samples for different modules can share source trajectories and execution prefixes, but not loss targets. Single LoRA follows the same recipe with rank 128, three epochs, and a peak learning rate of $3\times10^{-4}$.

\textbf{Full SFT.} The full-parameter baseline starts from the same Qwen3-32B checkpoint and is trained with LLaMA-Factory on eight NVIDIA H200 GPUs, using DeepSpeed ZeRO-3 without offloading, FlashAttention-2, Liger kernels, gradient checkpointing, and bfloat16. Each optimizer step uses 64 packed sequences: one per device per micro-step, with eight gradient-accumulation steps. AdamW uses a peak learning rate of $4\times10^{-5}$, warmup over the first 10\% of steps followed by cosine decay, and no weight decay. Training runs for five epochs with seed 42. Trajectories are packed into 32k token sequences with block-diagonal attention, so packed trajectories do not attend to one another. The Qwen3 template keeps the reasoning content, and prompt and tool-response tokens are excluded from the loss. Full SFT trains on compressed trajectories (Appendix~\ref{app:segmentation}).

\textbf{Evaluation protocol.} The main table reports each benchmark separately, and its Avg. column is the unweighted mean over the eight benchmarks. Each method keeps its own context management and model-call structure and attempts every task once. A task that ends without a valid boxed answer is retried up to twice, allowing at most three attempts; the first valid answer is scored. On 2Wiki, Bamboogle, and MuSiQue, an answer is correct if the boxed answer exactly matches the gold answer or one of its aliases after normalization (lowercasing and removing punctuation, articles, and extra whitespace). On the other five benchmarks, GPT-5.6 Luna compares the answer with the reference using each benchmark's grading prompt; on GAIA, WebWalkerQA, and FRAMES, answers that match the reference after normalization are accepted without a judge call. For a benchmark with accuracy $p$ on $n$ evaluation tasks, Table~\ref{tab:main-results} gives the nominal task-based binomial standard error $\sqrt{p(1-p)/n}$. For Avg., we propagate these as $\sqrt{\sum_b \mathrm{SE}_b^2}/8$, treating benchmarks as independent.

\section{Trace Analysis Details}
\label{app:trace-analysis}

\begin{table}[!t]
\centering
\begingroup
\fontsize{10}{12}\selectfont
\setlength{\tabcolsep}{6pt}
\begin{tabular}{@{}lccc@{}}
\toprule
Method & QU$\uparrow$ & Cov$\uparrow$ & PK$\downarrow$ \\
\midrule
ReAct & 0.365 & 0.336 & 0.590 \\
Least-to-Most & 0.386 & 0.360 & 0.600 \\
ManuSearch & 0.305 & 0.385 & 0.520 \\
\midrule
Single LoRA & 0.315 & 0.332 & 0.630 \\
Full SFT & 0.410 & 0.478 & 0.530 \\
\midrule
TIES-R64 & 0.293 & 0.311 & 0.680 \\
DARE-R64 & 0.370 & 0.375 & 0.600 \\
TSPA-R64 & 0.375 & 0.343 & 0.640 \\
\midrule
\method-hard & \textbf{0.570} & \textbf{0.601} & \textbf{0.390} \\
\method-soft & 0.521 & 0.554 & 0.440 \\
\bottomrule
\end{tabular}
\endgroup
\caption{SearchAtlas diagnostics by method over the 100 tasks of Figure~\ref{fig:trace-dag}. Values are task means; QU = query utilization, Cov = question-constraint coverage (answer graph), PK = share of answers with a prior-knowledge-to-answer edge. Bold marks the best value per column (lowest for PK).}
\label{tab:trace-dag-metrics}
\end{table}

\textbf{Graph construction.} We use SearchAtlas's released builder and evaluator to represent searches as query nodes, with page reads supplying evidence to the queries that retrieved them. Edges link queries, question constraints, prior knowledge, and the final answer. Attribution uses GPT-5.2 at temperature 0 with JSON outputs and access to all earlier tool results. Each question is decomposed once into constraint units shared across methods; evidence and answer attribution are performed separately for each method. Reader summaries are treated as observed content without checking the original pages.

\textbf{Diagnostics.} Let $\mathcal{Q}$ be the issued queries, $A$ the answer, and $\mathcal{Q}_A\subseteq\mathcal{Q}$ the queries that reach $A$ through evidence-use edges. Let $C$ be the question's constraint units, $C(q)\subseteq C$ those deployed by query $q$, and $E_{\mathrm{PK}}$ the prior-knowledge edges. For nonempty query and constraint sets, the diagnostics used here are
\begin{gather}
\mathrm{QU} = \frac{|\mathcal{Q}_A|}{|\mathcal{Q}|}, \\
\mathrm{Cov} = \frac{\left|\bigcup_{q\in\mathcal{Q}_A} C(q)\right|}{|C|}, \\
\mathrm{PKA}_{\mathrm{ans}} = \mathbf{1}\!\left[(\mathrm{PK},A)\in E_{\mathrm{PK}}\right].
\end{gather}
QU measures the fraction of queries on an evidence path to the answer. Cov uses the constraints deployed by direct answer parents and their evidence ancestors, consistently across tasks. PKA$_{\mathrm{ans}}$ indicates an answer edge attributed to prior knowledge without retrieved support. Reported scores are task means, so mean PKA$_{\mathrm{ans}}$ is the share of answers with such an edge. Higher QU and Cov and lower PKA$_{\mathrm{ans}}$ indicate stronger retrieval grounding. Low coverage may indicate missing retrieval support for some question requirements; unsupported content is not necessarily incorrect.

\textbf{Trajectories.} Except as noted in Section~\ref{sec:exp:trace-analysis}, we use for each method and task the trajectory behind the corresponding result in Table~\ref{tab:main-results}: the earliest successful attempt for correctly answered tasks and the last completed attempt for tasks answered incorrectly.

\textbf{Quality control and statistics.} A task is excluded for all methods if any method's graph contains CJK text (the constraint matching relies on English lexical signals), has no answer signals, or records an unrecovered constraint-arbitration error; excluded tasks are not replaced. The 100 retained tasks come from MuSiQue (23), 2Wiki (20), Bamboogle (18), GAIA (17), FRAMES (14), BrowseComp (4), and WebWalkerQA (4); xbench-DS tasks do not pass the language filter. Means weight tasks equally. Intervals are 95\% percentile intervals from 5{,}000 task-level bootstrap resamples.

\textbf{Additional observations.} Table~\ref{tab:trace-dag-metrics} reports the means in Figure~\ref{fig:trace-dag}. Both \method variants lead all baselines on the three diagnostics. ManuSearch issues the most queries, but \method-hard covers more question constraints along the evidence paths supporting its answers. \method-hard also has fewer tasks without a retrieval-supported answer path than any baseline.

\begin{table}[h]
\centering
\small
\setlength{\tabcolsep}{4.0pt}
\begin{tabular}{@{}lcccc@{}}
\toprule
 & MegaScience & SciBench & DS-1000 & GAIA \\
\midrule
\#Tasks & 200 & 575 & 100 & 103 \\
\bottomrule
\end{tabular}
\caption{The number of selected tasks used for evaluating code insertion tasks. The MegaScience and DS-1000 tasks were randomly sampled based on a balanced consideration of cost and representativeness.}
\label{tab:code-extension-sets}
\end{table}

\section{General Capability Preservation}
\label{app:preservation}
\label{sec:exp:preservation}

We evaluate catastrophic forgetting on \dataset{MMLU-Pro}~\citep{mmlupro}, \dataset{IFEval}~\citep{ifeval}, \dataset{MT-Bench}~\citep{mtbench}, \dataset{StructFlow}, and \dataset{HumanEval}. All external tools are disabled: \method reasons through model weights while retaining internal adapter selection.

Table~\ref{tab:general-capability-scores} shows that \method substantially mitigates the general-capability forgetting observed under Full SFT across these five benchmarks. Full SFT loses 11.60 and 15.16 points on MMLU-Pro and IFEval, respectively. Both \method variants exceed Base on MMLU-Pro and MT-Bench and remain close to it on StructFlow. IFEval still declines, but much less than under Full SFT; HumanEval matches Base under soft routing and declines less than Full SFT under hard routing. Both variants thus retain higher scores than Full SFT on every evaluated benchmark.

\begin{remark}
\label{rem:zero-drift}
With every adapter inactive, the model parameters are bitwise equal to the frozen base $\theta_0$; with adapter $k$ active, they differ by $\Delta\theta_k$. This concerns parameters only; Table~\ref{tab:general-capability-scores} measures behavior with adapters available.
\end{remark}

\begin{table}[t]
\centering
\begingroup
\fontsize{10}{12}\selectfont
\setlength{\tabcolsep}{1pt}
\definecolor{oursband}{RGB}{228,243,255}
\definecolor{retentionbest}{RGB}{0,100,0}
\definecolor{retentionworst}{RGB}{200,0,0}
% Natural column widths and ordinary padding keep the blue band continuous.
\begin{tabular}{@{}lrrrrr@{}}
\toprule
\makecell[c]{Method} & \makecell[c]{MMLU-\\Pro} & \makecell[c]{IFEval} & \makecell[c]{MT-\\Bench} & \makecell[c]{Struct\\Flow} & \makecell[c]{Human\\Eval} \\
\midrule
Base & 58.03 & 77.82 & 9.3875 & 98.12 & 90.85 \\
\midrule
Full SFT & \textcolor{retentionworst}{$-19.99$} & \textcolor{retentionworst}{$-19.48$} & \textcolor{retentionworst}{$-0.93$} & \textcolor{retentionworst}{$-1.60$} & \textcolor{retentionworst}{$-6.86$} \\
DARE-TIES & $-0.55$ & $-4.61$ & $+0.53$ & $-0.95$ & $-1.34$ \\
Single LoRA & $-1.15$ & \textcolor{retentionbest}{$\boldsymbol{-3.80}$} & $0.00$ & $-0.76$ & $-5.37$ \\
\rowcolor{oursband}
\method-hard & \textcolor{retentionbest}{$\boldsymbol{+1.60}$} & $-4.51$ & \textcolor{retentionbest}{$\boldsymbol{+0.80}$} & $-0.08$ & $-2.69$ \\
\rowcolor{oursband}
\method-soft & $+1.21$ & $-4.28$ & \textcolor{retentionbest}{$\boldsymbol{+0.80}$} & \textcolor{retentionbest}{$\boldsymbol{-0.04}$} & \textcolor{retentionbest}{$\boldsymbol{0.00}$} \\
\bottomrule
\end{tabular}
\endgroup
\caption{General-capability retention (external tools disabled). Base: scores; others: $100(S/S_{\mathrm{Base}}-1)\%$. Green/red mark the best/worst changes per column, including ties.}
\label{tab:general-capability-scores}
\end{table}

% Queue the final wide floats early so they can share pages with their text.
% Optional appendix table; requires booktabs.
\begin{table*}[t]
\centering
\fontsize{10}{12}\selectfont
\setlength{\tabcolsep}{4pt}
\begin{tabular}{ccccccc}
\toprule
Rubric & \multicolumn{3}{c}{\method{}-Hard} & \multicolumn{3}{c}{\method{}-Soft} \\
\cmidrule(lr){2-4}\cmidrule(lr){5-7}
& Original & \writingplus{} & $\Delta$ & Original & \writingplus{} & $\Delta$ \\
\midrule
Comprehensiveness & 25.32 & 26.33 & $+1.01$ & 26.50 & 26.78 & $+0.28$ \\
Depth & 22.46 & 23.59 & $+1.13$ & 23.82 & 24.19 & $+0.37$ \\
Instruction following & 29.65 & 31.22 & $+1.57$ & 31.36 & 32.45 & $+1.09$ \\
Readability & 32.33 & 33.81 & $+1.48$ & 33.32 & 35.10 & $+1.77$ \\
\bottomrule
\end{tabular}
\caption{DeepResearchBench rubric scores before and after the Writing refresh (higher is better). $\Delta$ is the reported score-point change from Original to \writingplus{}.}
\label{tab:writing-refresh-rubrics}
\end{table*}

% Use the full text width to retain readable probe columns.
\begin{table*}[tp]
\centering

\fontsize{10}{12}\selectfont
\setlength{\tabcolsep}{5pt}
\begin{tabular}{cccccc}
\toprule
Evaluation Set & \multirow{2}{*}[-\dimexpr(\aboverulesep+\belowrulesep+\cmidrulewidth)/2\relax]{Rows} & \multicolumn{2}{c}{NLL} & \multicolumn{2}{c}{Perplexity} \\
\cmidrule(lr){3-4}\cmidrule(lr){5-6}
& & Original & \writingplus{} & Original & \writingplus{} \\
\midrule
OpenScholar validation & 205 & 1.27 & 1.02 & 3.56 & 2.76 \\
Historical Writing validation & 200 & 0.46 & 0.50 & 1.59 & 1.66 \\
\bottomrule
\end{tabular}
\caption{Supervised-token fit before and after the Writing refresh (lower is better). }
\label{tab:writing-refresh-loss}
\end{table*}

\section{Code Insertion: Implementation and Evaluation Protocol}
\label{app:code-extension}

Insertion adds a \texttt{code} capability without retraining the base, modifying the \texttt{plan}, \texttt{search}, and \texttt{writing} adapters, or replaying their data. We train only a new \texttt{code} adapter on code-computation trajectories from MiroVerse-MegaScience, converted to our tool and handoff protocol; these trajectories use code to calculate and verify answers from given conditions. Registering the \texttt{code} alias extends the routing vocabulary from $K=3$ to $K=4$.

The evaluation has two parts. First, we compare hashes of the base and the three existing adapters before and after insertion. Second, as shown in Table~\ref{tab:code-extension-sets}, we evaluate the original \dataset{GAIA} tasks before and after insertion, reporting the paired quality change and how often \texttt{code} is invoked on tasks that need no computation (2 of 80). New-capability utility is measured on \dataset{MegaScience}~\citep{megascience}, \dataset{SciBench}~\citep{scibench}, and \dataset{DS-1000}~\citep{ds1000}. Insertion succeeds only if code tasks improve, the existing hashes are unchanged, old-task quality does not decrease, and misrouting into \texttt{code} stays low; unchanged parameters alone do not suffice.

% Code insertion expands the capability library with stable performance.

\section{Writing Refresh: Protocol and Full Results}
\label{app:writing-refresh}

\textbf{Local update.} We continue training Writing on OpenScholar-based trajectories synthesized in our harness, keeping the base, Plan, and Search fixed and replaying no earlier data. The pool contains 1,000 rows from 983 unique trajectories. Splitting by trajectory ID gives 795 training rows and 205 validation rows; only the training split is used for the update.

\textbf{Report-quality results.} Table~\ref{tab:writing-refresh-rubrics} gives the DRB rubric breakdown for the overall scores in Table~\ref{tab:writing-refresh-scores}. DRB totals use the evaluator's weighted aggregation, not an unweighted rubric mean.

\textbf{Auxiliary adaptation and retention probe.} The paired teacher-forced probe uses identical reference sequences, tokenization, and masks for the original and refreshed Writing adapters. It activates Writing on Writing positions and uses the frozen base elsewhere. NLL is averaged over supervised tokens and exponentiated for perplexity; targets may include reasoning and tool calls as well as report text. Table~\ref{tab:writing-refresh-loss} reports NLL and perplexity for the OpenScholar and historical Writing validation sets under this shared protocol.

\end{document}